\documentclass{article}

\usepackage{arxiv}

\usepackage[utf8]{inputenc} 
\usepackage[T1]{fontenc}    
\usepackage{hyperref}       
\usepackage{url}            
\usepackage{booktabs}       
\usepackage{amsfonts}       
\usepackage{nicefrac}       
\usepackage{microtype}      
\usepackage{lipsum}
\usepackage{amsmath}
\usepackage{pgfplots}
\usepackage{subfigure}

\usepackage{hyperref}
\hypersetup{
    colorlinks=true,
    linkcolor=blue,
    urlcolor=red,
    pdftitle={Research Statement}
    }
\usepackage{fancyhdr}
\usepackage{setspace}
\usepackage{amsmath}
\usepackage{amssymb}
\usepackage{makeidx}
\usepackage{lmodern}
\usepackage{fourier}
\title{Analyzing the Performance of Deep Encoder-Decoder Networks as Surrogates for a Diffusion Equation}

\author{
  J. Quetzalc\'oatl Toledo-Mar\'in \\
  University of British Columbia, \\
  BC Children's Hospital Research Institute \\
  Vancouver BC, Canada \\
  \texttt{j.toledo.mx@gmail.com} \\
  \And
  James A. Glazier \\
  Biocomplexity Institute and \\
  Department of Intelligent Systems Engineering, \\
  Indiana University, Bloomington, IN 47408, USA \\
  \texttt{jaglazier@gmail.com} \\
    \And
 Geoffrey Fox \\
 University of Virginia, \\ 
 Computer Science and Biocomplexity Institute, \\
 994 Research Park Blvd, Charlottesville, \\
 Virginia, 22911, USA\\
  \texttt{gcfexchange@gmail.com} \\
}
\date{}

\begin{document}
\maketitle

\begin{abstract}

Neural networks (NNs) have been demonstrated to be a viable alternative to traditional direct numerical evaluation algorithms, with the potential to accelerate computational time by several orders of magnitude. In the present paper we study the use of encoder-decoder convolutional neural network (CNN) algorithms as surrogates for steady-state diffusion solvers. The construction of such surrogates requires the selection of an appropriate task, network architecture, training set structure and size, loss function, and training algorithm hyperparameters. It is well known that each of these factors can have a significant impact on the performance of the resultant model. Our approach employs an encoder-decoder CNN architecture, which we posit is particularly well-suited for this task due to its ability to effectively transform data, as opposed to merely compressing it. We systematically evaluate a range of loss functions, hyperparameters, and training set sizes. Our results indicate that increasing the size of the training set has a substantial effect on reducing performance fluctuations and overall error. Additionally, we observe that the performance of the model exhibits a logarithmic dependence on the training set size. Furthermore, we investigate the effect on model performance by using different subsets of data with varying features. Our results highlight the importance of sampling the configurational space in an optimal manner, as this can have a significant impact on the performance of the model and the required training time. In conclusion, our results suggest that training a model with a pre-determined error performance bound is not a viable approach, as it does not guarantee that edge cases with errors larger than the bound do not exist. Furthermore, as most surrogate tasks involve a high dimensional landscape, an ever increasing training set size is, in principle, needed, however it is not a practical solution.
\end{abstract}




\section{Introduction}
Diffusion is ubiquitous in physical, biological and engineered systems. In mechanistic computer simulations of the dynamics of such systems, solving the steady state and time-varying diffusion equations with multiple sources and sinks is often the most computationally expensive part of the calculation, especially in cases with multiple diffusing species with diffusion constants differing by multiple orders of magnitude. 
In real-world problems, the number of sources and sinks, their shape,  boundary fluxes and positions differ from instance to instance and may change in time. Boundary conditions may also be complicated and diffusion constants may be anisotropic or vary in space. The resulting lack of symmetry means that many high-speed implicit and frequency-domain diffusion-solver approaches do not work effectively, requiring the use of simpler but slower forward solvers \cite{schiesser2012numerical}. Deep learning surrogates to solve either the steady-state field or the time-dependent field for a given set of sources and sinks subject to diffusion can increase the speed of such simulations by several orders of magnitude compared to the use of direct numerical solvers. 

Deep learning techniques have been successfully applied to a wide range of problems by means of solving the corresponding sets of partial differential equations. This has been extensively studied in literature, with references such as \cite{farimani2017deep, sharma2018weakly, he2020unsupervised, edalatifar2020using, li2020reaction, li2020fourier, cai2021physics}. See Ref. \cite{fox2019learning} for a thorough review. Furthermore, embedding neural networks in differential equations has also shown promising results in the area of Neural-ODEs.
Modeling a physical process via an ordinary differential equation (ODE) typically involves equating the rate of change of a quantity, such as a concentration, to an operator applied to that quantity and any additional quantities, known as inhomogeneities or external fields. The ODE is then solved and compared to experimental data to validate the model or fit parameters. The operator in the ODE is traditionally chosen based on the symmetries of the physical process. However, in the case of Neural-ODEs, the operator is replaced with a neural network, which is trained by solving the Neural-ODE and comparing it to experimental data \cite{chen2018neural, DBLP:journals/corr/abs-1902-02376}. Diffusion models are an example of Neural-ODEs  \cite{song2020score}.
Additionally, Physics-Informed Neural Networks (PINNs) have been proposed as a means of addressing forward and inverse problems by embedding physical information into the neural network. This can be achieved by embedding the ODE, initial conditions and boundary conditions into the loss function used to train the neural network \cite{raissi2019physics}. These methods have been successful in various fields, such as computational molecular dynamics \cite{noe2020machine, majewski2022machine, entwistle2022electronic}. However, the complexity of multiscale systems, which often exhibit different characteristic length and time scales differing by orders of magnitude, in addition to the lack of symmetries, can make the solution of these systems challenging.
AI-based surrogates using deep learning methods can be used to accelerate computation by replacing specific classical solvers, while preserving the interpretability of mechanistic models. In the field of multiscale modeling, AI-based surrogates can provide a computationally efficient alternative to traditional approaches such as Monte Carlo methods and molecular dynamics.

\begin{figure}[hbtp]
\centering
\subfigure[Initial condition]{
\includegraphics[width=2.0in]{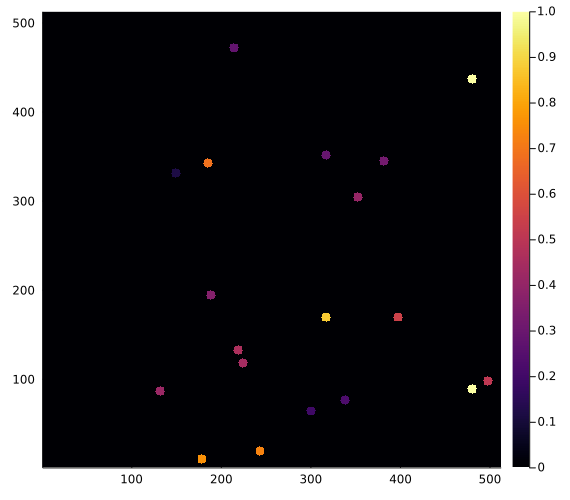}
\label{fig:in}}
\centering
\subfigure[Stationary solution]{
\includegraphics[width=2.0in]{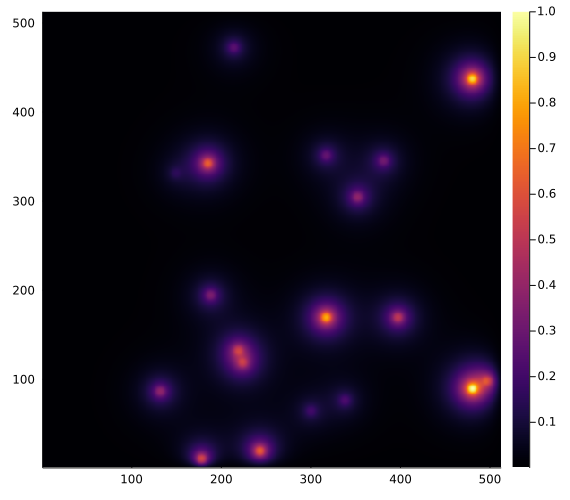}
 \label{fig:outSS}}
 \subfigure[Prediction]{
\includegraphics[width=2.0in]{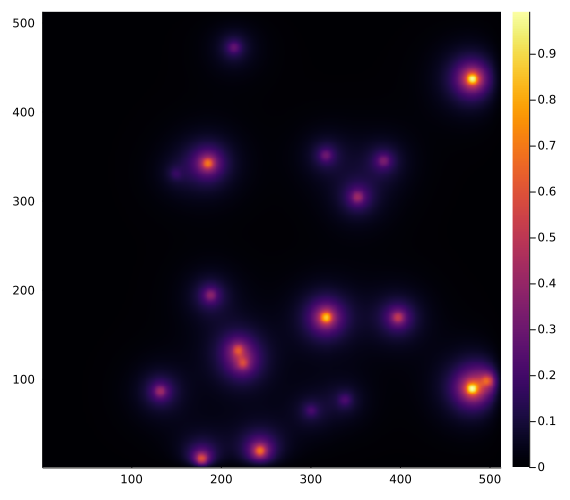}
 \label{fig:out}}
\caption{ Snapshot of \textbf{a)} initial condition, \textbf{b)} stationary state solution and \textbf{c)} the prediction of one of our trained models. \textbf{a)} We placed 19 random value sources of radius $5\;voxels$ in random positions fully within a $512\times512 pixel$ lattice and used this configuration as the input to the NN. \textbf{b)} Stationary solution to the diffusion equation with absorbing boundary conditions for the initial conditions in \textbf{a)} where the sources have a constant random flux. The stationary solution \textbf{b)} is the target for the NN, while the prediction \textbf{c)} is the output of the NN given \textbf{a)} as input. We fixed the diffusion constant to $D = 1\;voxels^2/s$ and the decay rate to $\gamma = 1/400s^{-1}$, which yields a diffusion length equal to $\sqrt{D/\gamma}\;voxels = 20 voxels$.
} \label{fig:inout}
\end{figure}


The development of effective deep neural network (NN) surrogates for diffusion-solver problems is a challenging task. This is due to several factors, such as the high dimensionality of the problem specification, which includes an arbitrary pattern of sources and sinks, varying boundary conditions for each source and sink, and spatially variable or anisotropic diffusivities. Additionally, there are a number of design decisions that must be made, such as the selection of the network architecture, the structure and size of the training set, the choice of loss function, and the selection of hyperparameters for training algorithms. Furthermore, it is essential to define accurate metrics for evaluating the task-specific performance of the trained network, as this may differ from the loss function used during training.


In prior work \cite{toledo2021deep}, we demonstrated the utility of deep convolutional neural networks in predicting the stationary solution for a diffusion problem on a $100\times100$ lattice with two circular random sources of radius $5$ units and random fixed flux, located in random positions, under absorbing boundary conditions. Here, we extend this approach by increasing the number of sources and the lattice size to $512\times512$.
The problem specification is as follows: we assume absorbing boundary conditions and the sources are fully contained within the domain. Each source imposes a constant value on the diffusing field within the source and at its boundary, with one of the sources having a value of 1 and the other sources having values randomly selected from a uniform distribution between $(0,1]$ (as depicted in Figure \ref{fig:in}). Outside of the sources, the field diffuses with a constant diffusion constant ($D$) and linearly decays with a constant decay rate ($\gamma$). This simple geometry could represent the diffusion and uptake of oxygen in a volume of tissue between parallel blood vessels of different diameters.
Although reflecting or periodic boundary conditions may better represent a portion of a larger tissue, we use the simpler absorbing boundary conditions here. The steady-state field in this case is highly dependent on the distance between the sources, the distance between the sources and the boundary, both relative to the diffusion length ($l_{D} = (D/\gamma)^{1/2}$) and on the sources' field strengths.


In practice, the steady-state diffusion equation maps an input image consisting of $512 \times 512$ pixels with a value of 0 outside the sources and constant values between 0 and 1 inside the sources, to an output image of the same size, which is a superposition of exponentially decaying functions, with its maximum at the center of each source (as illustrated in Figure \ref{fig:out}). In order to evaluate the performance of a neural network (NN) in approximating the steady-state diffusion field for configurations of sources that it had not previously encountered, we trained the NN on explicit numerical solutions of the steady-state diffusion field for $16,000$ examples per number of sources.

It is well-established that the diffusion kernel convolution used in the direct solution of the time-dependent diffusion equation, such as finite-element methods, is a type of convolutional neural network (CNN) \cite{schiesser2012numerical}. Therefore, in our study, we chose to utilize deep convolutional neural networks (DCNNs) as the architecture. However, there are various types of CNNs that can be employed. In this study, we considered a deep convolutional neural network with the archetype of an encoder-decoder \cite{baur2020autoencoders}. We argue that the encoder-decoder CNN architecture is particularly well-suited for this task, as it effectively transforms data, rather than compressing it, akin to the Fourier transform. Additionally, we evaluated different loss functions and prefactors to weigh in the source regions in the lattice, which are statistically undersampled in the training set. We also assessed and discussed the impact of different hyperparameters and training set sizes on the model performance.

However, in order to accurately evaluate the performance, it is essential to define a suitable metric. We discussed different metrics to evaluate the different models trained. Our results suggest that increasing the size of the training set has the most significant effect on reducing performance fluctuations and overall error. Furthermore, we studied the performance of models versus training set size and found that as the training set size increases, the performance fluctuations decrease, and the computational time for training increases linearly. In addition, the model performance has a logarithmic dependence on the training set size.

It is crucial to acknowledge that our findings do not indicate the existence of an optimal training set size. Rather, our results suggest that training a model with a pre-determined error performance bound is not a viable approach, as it does not guarantee that edge cases with errors larger than the bound do not exist. Furthermore, as most surrogate tasks involve a high dimensional landscape, an ever increasing training set size is, in principle, needed, however it is not a practical solution.


\section{Methods}
The steady state of the diffusion equation is precisely the convolution of the equation's associated Green's function and the initial state. Hence a naive approach would consider a deep convolutional NN (CNN) where the height and width in each layer is kept fixed and equal to the input. CNNs are good at detecting edges and as the number of convolution layers increases, the CNN can detect more sophisticated elements. So can a CNN predict the stationary solution? We argue that it would require a large number of convolution layers such that each convolution would be proportional to a timestep. Adding many convolutions would lead to a rather harder and slower training, as it would increase the chances of under or overflow in the gradient. Furthermore, the number of convolution layers would need to be variable is some form or fashion in order to mimic the timesteps required to reach the stationary solution given an initial condition. There are ways to address each of the previous setbacks but there's also a trade-off for every decision (\textit{e.g.}, increasing the number of convolutional layers increases the model's instability which can be addressed by adding more regularization layers which makes the model larger and requires more time to train). Experience tells us that deep convolutional NNs where the height and width in each layer is kept fixed and equal to the input are not good for our task \cite{toledo2021deep}.



\begin{figure}[hbtp]
\centering
\includegraphics[width=4.6in]{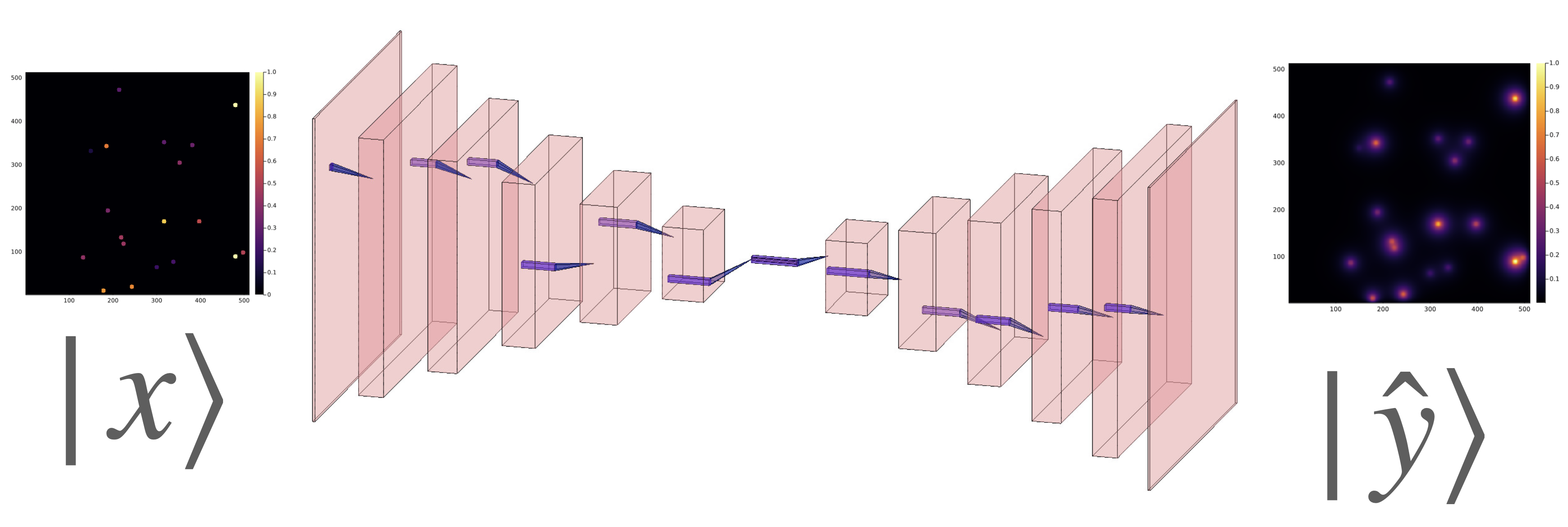}
 \label{fig:NN2}
\caption{ Sketch of the NN's architecture. The first 6 layers perform a meanpool operation that reduces the height and width by half after each layer, with the image dimensions following the sequence $\lbrace 512^2, 256^2, 128^2, 64^2, 32^2, 16^2, 16^2 \rbrace$ while adding channels after each layer following the sequence $\lbrace 1,64,128,256,512,1024,2048 \rbrace$. Then, the following 7 blocks of layers reverse the process, reducing the number of channels following the sequence $\lbrace 1024, 512, 256, 128, 64, 32, 1 \rbrace$ while increasing the height and width following the sequence $\lbrace 16^2, 32^2, 64^2, 128^2, 256^2, 512^2, 512^2 \rbrace$.
This sketch only defines the kind of layers used. For details about the activation functions used in each layer, see Table \ref{table:1}.
} \label{fig:NN}
\end{figure}

We designed and built an encoder-decoder CNN (ED-CNN) as our architecture. Fig. \ref{fig:NN} provides a sketch of our NN architecture. We denote by $| x\rangle$ and $| \hat{y} \rangle$ the input and output images, that is the initial condition layout of the source cells and the predicted stationary solution of the diffusion equation, respectively. The input $| x\rangle$ goes through convolutional blocks depicted in Fig. \ref{fig:NN} and generates the prediction $| \hat{y} \rangle$ which we then compare with the ground truth $|y \rangle$. The core feature of an encoder-decoder is the cone-like structure whereby the information is being compressed when it reaches the bottle neck and then it's decompressed.

Consider the following: A CNN can be seen as a set of operators $\lbrace O_i \rbrace$ acting on the input, \textit{i.e.}, \begin{equation}
    | \hat{y} \rangle = \prod_{i=1}^n O_i | x \rangle \; , \label{eq:Ops}
\end{equation}
where the $i$ tags the convolutional layer. Notice that in the particular case where $\lbrace O_i \rbrace$ are unitary linear operators and all the same $O= \mathbf{1} - A dt$, Eq. \eqref{eq:Ops} becomes
\begin{equation}
    |\hat{y} \rangle = (\mathbf{1}-A dt)^n | x \rangle \; .
\end{equation}
Notice that the previous Eq. is akin to a ResNet architecture \cite{he2016deep}.
By defining $dt = t/n $, where $t$ is time, and taking the limit $n \rightarrow \infty $ leads to $| \hat{y} \rangle = \exp(A t) | x \rangle$. Hence, when $| \hat{y} \rangle$ equals the ground truth $| y \rangle$, the CNN has learnt the operator $\exp(At)$. Notice that taking the derivative with respect to time leads to the diffusion equation:
\begin{equation}
    \frac{\partial | \hat{y} \rangle}{\partial t} = A | \hat{y} \rangle \; .
\end{equation}


Typically, numerical methods for searching for the stationary solution of a diffusion equation operate by comparing the solution at time $t$ with the solution at time $t+\Delta t$, and halting the process when the difference between $|y(t) \rangle$ and $|y(t+\Delta t) \rangle$ is smaller in magnitude than some pre-determined error value. However, the time $t$ at which the stationary solution is reached depends on the initial conditions and the inhomogeneity term in the diffusion equation. This analogy suggests that, similar to traditional numerical methods, CNNs also require a flexible number of layers to reach the stationary solution, rather than an infinite number of layers. There have been recent developments in addressing the issue of a variable number of layers. For example, in reference \cite{kc2021joint}, the authors propose a method in which the dropout parameters are trained in conjunction with the model parameters.


In the context of ED-CNNs, it is a widely held belief that the cone-like architecture leads to improved performance due to the compression in the bottleneck. However, we argue that this is not the sole, if any, contributing factor. It is important to note that the ED-CNN architecture involves summing over the lattice and introducing a new variable, the depth. Additionally, past the bottleneck, the ED-CNN projects back to the original space.
In this sense, we say the ED-CNN learns a propagator in a reciprocal space similar to methods such as Laplace and Fourier transform. 
In Table \ref{table:1} we specify the neural network by specifying the type of operation and output shape through the sequence of blocks in the NN. Notice that the bottleneck block is $2048*16*16$ which is twice the size of the input.


\begin{table}[h!]
\centering
{\renewcommand{\arraystretch}{1.3} 
\begin{tabular}{|c c|} 
 \hline
 Operation & Output shape \\ [0.5ex] 
 \hline\hline
 Conv 3 x 3 & 64 x 512 x 512 \\ 
 \hline
 BatchNorm & - \\
 \hline
 LReLU(0.02) & - \\
 \hline
 AvgPool2d & 64 x 256 x 256 \\
 \hline
 Conv 3 x 3 & 128 x 256 x 256 \\ 
  \hline
 LReLU(0.02) & - \\
 \hline
 AvgPool2d & 128 x 128 x 128 \\
 \hline
 Conv 3 x 3 & 256 x 128 x 128 \\ 
 \hline
 LReLU(0.02) & - \\
 \hline
 AvgPool2d & 256 x 64 x 64 \\
 \hline
 Conv 3 x 3 & 512 x 64 x 64 \\ 
 \hline
 LReLU(0.02) & - \\
 \hline
 AvgPool2d & 512 x 32 x 32 \\
 \hline
 Conv 3 x 3 & 1024 x 32 x 32 \\ 
 \hline
 LReLU(0.02) & - \\
 \hline
 AvgPool2d & 1024 x 16 x 16 \\
 \hline
 Conv 3 x 3 & 2048 x 16 x 16 \\ 
 \hline
 LReLU(0.02) & - \\
 \hline
 \end{tabular}
 \begin{tabular}{|c c|} 
 \hline
 Operation & Output shape \\ [0.5ex] 
 \hline\hline
 ConvT 3 x 3 & 1024 x 16 x 16 \\ 
 \hline
 LReLU(0.02) & - \\
 \hline
 ConvT 3 x 3 & 512 x 16 x 16 \\ 
 \hline
 LReLU(0.02) & - \\
 \hline
 Upsample & 512 x 32 x 32   \\
 \hline
 ConvT 3 x 3 & 256 x 32 x 32 \\ 
 \hline
 LReLU(0.02) & - \\
 \hline
 Upsample & 256 x 64 x 64  \\
 \hline
 ConvT 3 x 3 & 128 x 64 x 64 \\ 
 \hline
 LReLU(0.02) & - \\
 \hline
 Upsample & 128 x 128 x 128  \\ 
 \hline
 ConvT 3 x 3 & 64 x 128 x 128 \\ 
 \hline
 LReLU(0.02) & - \\
 \hline
 Upsample & 64 x 256 x 256  \\ 
 \hline
 ConvT 3 x 3 & 32 x 256 x 256 \\ 
 \hline
 LReLU(0.02) & - \\
 \hline
 Upsample & 32 x 512 x 512  \\ 
 \hline
 ConvT 4 x 4 & 1 x 512 x 512 \\ 
 \hline
 LReLU(0.02) & - \\
 \hline
\end{tabular}

\caption{Convolutional Neural Network architecture. \textit{Act} stands for activation function. Conv, ConvT and (L)ReLU stand for convolution, convolution transpose, and (leaky) rectified linear unit, while Identity means the activation function is the identity function (see Ref. \cite{Flux.jl-2018}). The NN take as input the initial condition which has dimensions $Channels \times Width \times Height = 1 \times 512 \times 512$ }
\label{table:1}
} 
\end{table}

To generate representative $\sigma$-source initial conditions and paired steady-state diffusion fields, we considered a two-dimensional lattice of size $512\times512\;\text{units}^2$. We generated 400k initial configurations with one to 20 sources randomly placed in the lattice, such that there are 20k per fixed number of sources (see Fig. \ref{fig:DSDiag}). Each source has a $5$-unit radius and for any given configuration one source has a constant flux value equal to $1$, while the other sources have a constant flux value between 0 and 1 randomly assigned using a uniform distribution.
Everywhere else the field value is zero. Each initial configuration served as the input for the NN $| x\rangle$. Then we calculated the stationary solution to the diffusion equation with absorbing boundary conditions for each initial condition using the \textit{Differential Equation} package in Julia \cite{rackauckas2017differentialequations}.
The Julia-calculated stationary solution is the target or ground truth image $ | y \rangle$ for the NN. In Figs. \ref{fig:in} and \ref{fig:out} we show an initial condition and the stationary solution, respectively.  
We set the diffusion constant to $D = 1\text{units}^2/\text{s}$ and the decay rate $\gamma = 1/400\text{s}^{-1}$, which yield a diffusion length $l_D = \sqrt{D/\gamma}=20\; \text{units}$. Notice that this length is 4 times the radius of the sources and $\sim 1/25$ the lattice linear dimension. As $\gamma$ increases and as $D$ decreases, this length decreases. As this length decreases, the field gradient also decreases \cite{tikhonov2013equations}. The source code to generate the data and train the NN can be found in Ref. \cite{githubJaque}.

\begin{figure}[hbtp]
\centering
\subfigure[Field]{
\includegraphics[height=2.0in,width=2.0in]{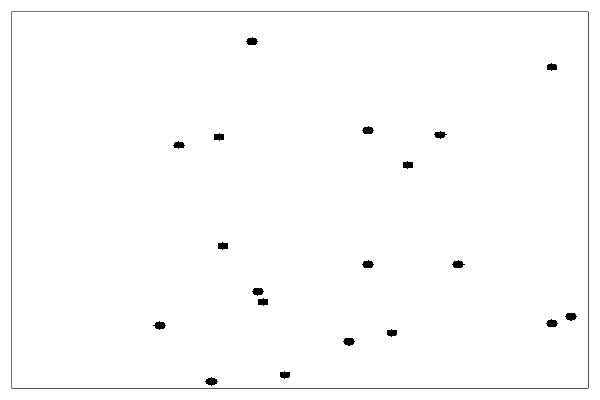}
\label{fig:fieldSamp}}
\centering
\subfigure[Sources]{
\includegraphics[height=2.0in,width=2.0in]{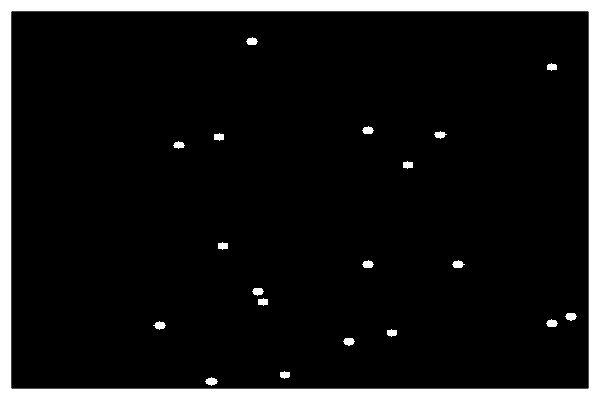}
 \label{fig:srcs}}
 \subfigure[Region 1]{
\includegraphics[height=2.0in,width=2.0in]{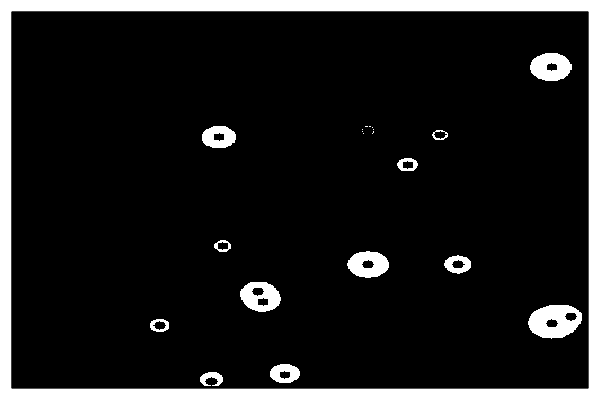}
 \label{fig:ring1}}
 \subfigure[Region 2]{
\includegraphics[height=2.0in,width=2.0in]{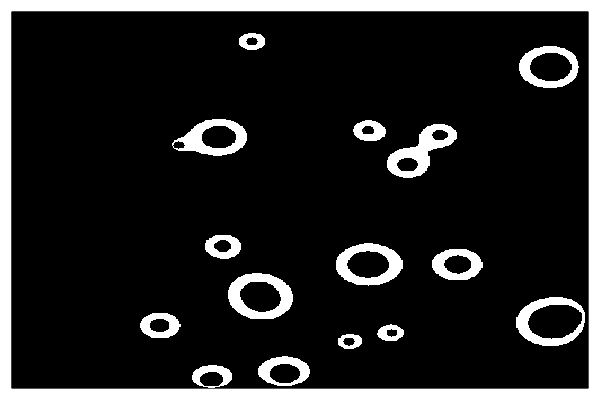}
 \label{fig:ring2}}
 \subfigure[Region 3]{
\includegraphics[height=2.0in,width=2.0in]{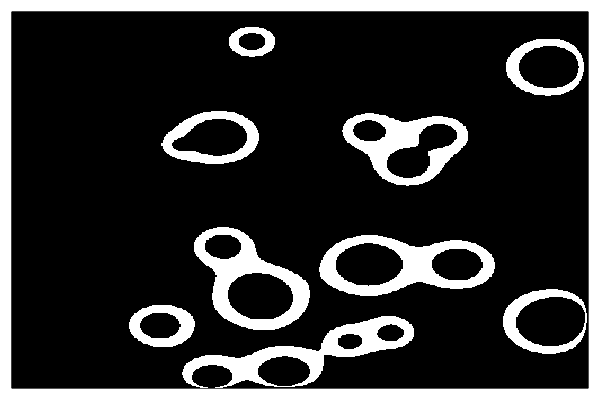}
 \label{fig:ring3}}
 \\
 \subfigure[Initial state]{
\includegraphics[width=2.0in]{Figs/input.png}
\label{fig:input-2}}
\subfigure[Stationary Solution]{
\includegraphics[width=2.0in]{Figs/field.png}
\label{fig:field-2}}
\subfigure[Average region coverage]{
\includegraphics[width=2.0in]{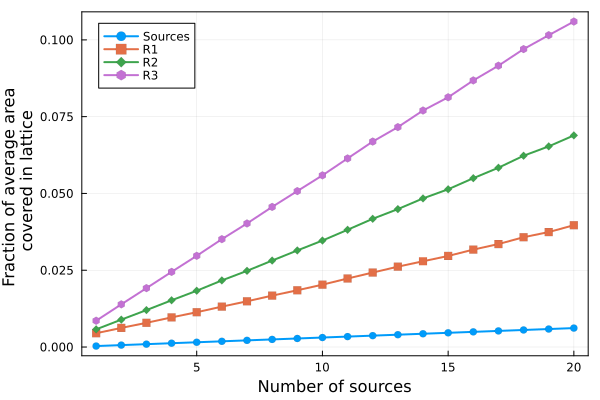}
\label{fig:area_cov}}
\caption{ Shown in white the \textbf{a)} field, \textbf{b)} sources, \textbf{c)} region 1, \textbf{d)} region 2 and \textbf{e)} region 3 regions in lattice given \textbf{f)} the input and \textbf{g)} the stationary state solution for 19 randomly placed sources. \textbf{h)} Average area fraction in lattice covered by R1,R2,R3 and sources \textit{vs} number of sources.} \label{fig:inout0}
\end{figure}

We built and trained the NN in PyTorch. We used ADAM as the optimizer \cite{kingma2014adam}. Deciding on a loss function is a critical choice in the creation of the surrogate. The loss function determines the types of error the surrogate's approximation will make compared to the direct calculation and the acceptability of these errors will depend on the specific application. The mean squared error (\textit{MSE}) error is a standard choice. However, it is more sensitive to larger absolute errors and therefore tolerates large relative errors at pixels with small values. 
In most biological contexts we want to have a small absolute error for small values and a small relative error for large values. We explored the NN's performance when trained using a set of different loss functions with different hyperparameters. The set of different loss functions we used were \textit{MAE}, \textit{MSE}, Huber loss \cite{huber1992robust} and inverse Huber loss defined as:
\begin{equation}
    IH_\delta(x) = 
    \begin{cases}
        \frac{1}{2}x^2, \qquad |x| > \delta \; , \\
        \delta (x - \frac{1}{2} \delta), \qquad |x| \leq \delta
    \end{cases} \; .
\end{equation}
The highest and lowest values in the input and output images are $1$ and $0$, respectively. The former only occurs in sources and their vicinity. Given the configurations of the sources, the fraction of pixels in the image with values near $1$ is $\sim \pi R^2/L^2 \approx 0.03-0.6\%$, as shown in  Fig. \ref{fig:area_cov}. 
Thus, pixels with small values are much more common than pixels with large values, and because the loss function is an average over the field, high field values tend to get washed out.
To account for this unbalance between the frequency of occurrence of low and high values, we considered prefactors to assign different weights to the error per pixel. One of which was an exponential weight on the pixels. We modulate this exponential weight through a scalar hyperparameter $w (\geq 0)$, such that the $i$th lattice position's contribution to the loss function yields:  
\begin{equation}
\mathcal{L}_{E,i \beta}^{(\alpha)} = \exp(-(\langle i |\mathbf{1}\rangle- \langle i | y_{\beta} \rangle)/w) \cdot \left(\langle i | \hat{y}_{\beta} \rangle - \langle i | y_{\beta} \rangle \right)^\alpha  \; , \label{eq:Loss_weight}
\end{equation}
where $\alpha$ is $1$ or $2$ for MAE or MSE, respectively and $\beta$ tags the tuple in the data set (input and target). Here $\langle | \rangle$ denotes the inner product and $|i\rangle$ is a unitary vector with the same size as $|y_{\beta} \rangle$ with all components equal to zero except the element in position $i$ which is equal to one. $|\mathbf{1} \rangle$ is a vector with all components equal to 1 and with size equal to that of $|y_{\beta} \rangle$. Then $\langle i | y_{\beta} \rangle$ is a scalar corresponding to the pixel value at the $i$th position in $|y_{\beta} \rangle$, whereas $\langle i | \mathbf{1}\rangle=1$ for all $i$. Notice that high and low pixel values will have an exponential weight $\approx 1$  and $\approx \exp(-1/w)$, respectively. This implies that the error associated to high pixels will have a larger value than that for low pixels. Another prefactor we considered was a step-wise function, \textit{viz}.
\begin{equation}
\mathcal{L}_{S,i \beta}^{(\alpha)} = (1 + a \tanh(b \langle i |y_\beta\rangle)) \cdot \left(\langle i | \hat{y}_{\beta} \rangle - \langle i | y_{\beta} \rangle \right)^\alpha  \; , \label{eq:Loss_step}
\end{equation}
where $a(\geq 0)$ and $b(\geq 0)$ are hyperparameters. As metioned earlier, we also trained models by replacing the metric term (MSE or MAE) with the Huber loss or the inverse Huber loss with hyperparameter $\delta$. The loss function $\mathcal{L}_{\chi}^{(\alpha)}$ is the mean value over all pixels ($i$) and data set ($\beta$) with $\chi = \lbrace E,S \rbrace$ for exponential weight and step weight, respectively, \textit{i.e.},
\begin{equation}
\mathcal{L}_{\chi}^{(\alpha)} = \langle \mathcal{L}_{\chi,i \beta}^{(\alpha)} \rangle \; .
\end{equation}
Here $\langle \bullet \rangle$ denotes average over pixels and data set. Each model was trained up to $100$ epochs unless stated otherwise, and saved the model with the lowest loss function value over the corresponding test set. We used $80\%$ and $20\%$ of the data set for training and test sets, respectively (see Fig. \ref{fig:DSDiag}). We trained different models and studied the effects different loss functions, and different data set size and features have in the model performance. In the next section we describe these and present the results.

\section{Results}

\subsection{Performance for different loss functions and data set sizes}

In this section we present the performance results of each of the models we trained. To this end, we use the MAE between prediction and target for each model. In general, models with good performance may not necessarily perform well in predicting sources and/or the field close to sources. In addition to computing the MAE on the whole lattice, we also measured the MAE in different regions in the lattice as shown in Fig. \ref{fig:inout0}. Filtering field and sources can be done using the initial state, while regions \textit{R1}, \textit{R2} and \textit{R3} correspond to pixel values in $[ 0.2, 1 ]$, $[ 0.1, 0.2 ]$ and $[ 0.05, 0.1 ]$, respectively. We compare the effects of different loss functions and different data set sizes. We validate all of our models using the same test set composed by $80k$ tuples (\textit{i.e.}, input and target). 

\begin{figure}[hbtp]
\centering
\includegraphics[width=6.2in]{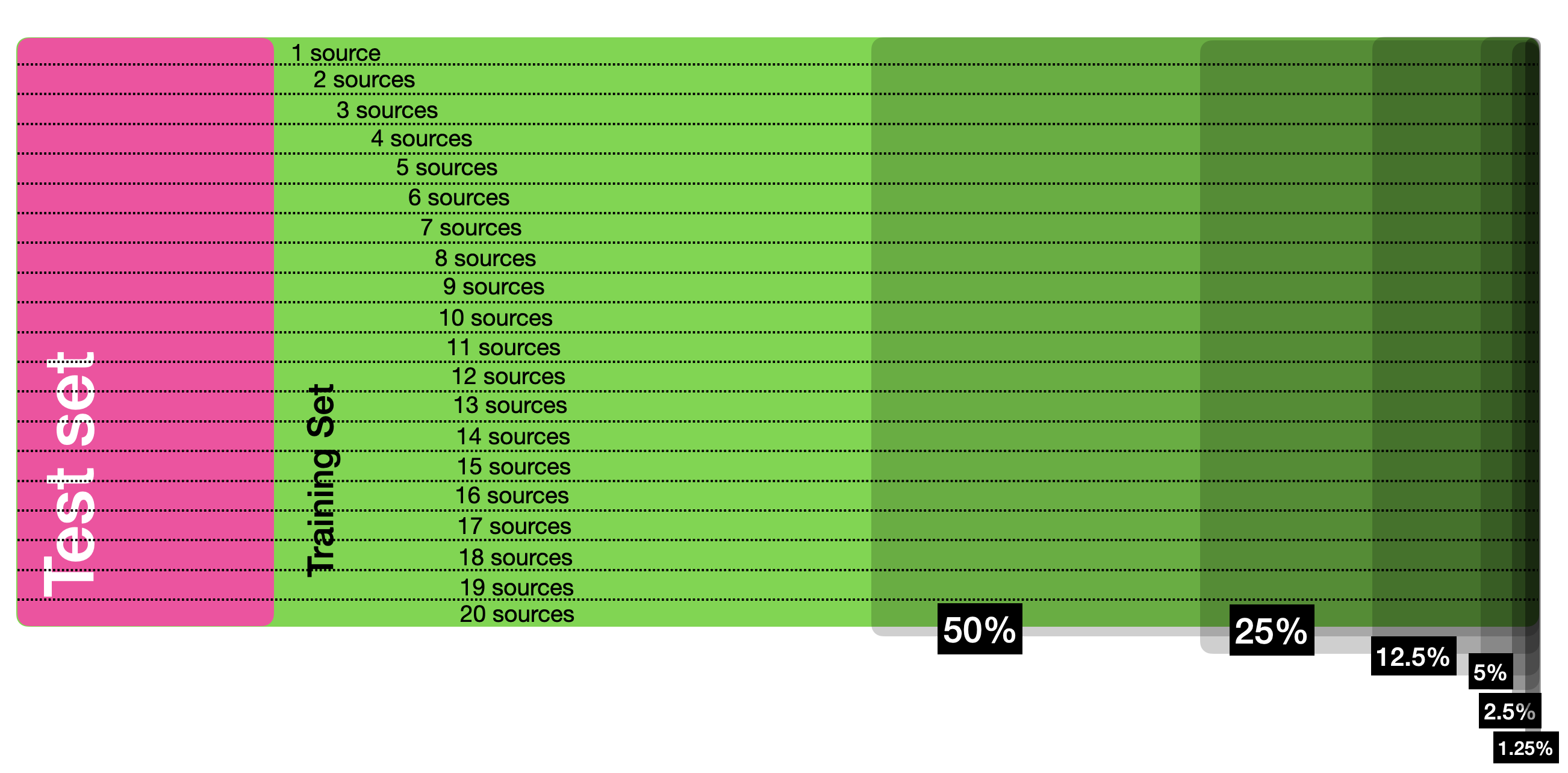}
\caption{Diagram of data set sizes. The data set is composed by 400k tuples (input and ground truth). We separate it in 20\% for test set (40k) and 80\% for training set (360k). We further take subsets $S_i$ of the training and test sets as depicted in the figure (50\%, 25\%, 12.5\%, 5\%, 2.5\% and 1.25\%). Each of these subsets contain equal number of samples with 1 to 20 sources. Each subset is contained in the immediate next, i.e., $S_{1.25\%} \subset S_{2.5\%} \subset S_{5\%} \subset S_{12.5\%} \subset S_{25\%} \subset S_{50\%} \subset S_{100\%}$. } \label{fig:DSDiag}
\end{figure}

In general, there are three main factors that contribute to the performance difference between different models, namely, the data set size, the loss function and the pseudo-stochasticity when training the model (weights initialization, minibatch parsing, etc). We found that the factor that contributes the most to performance is the data set size. For clarity purposes, we present our results by clustering models by data size trained on. We trained models using $1.25\%,2.5\%,5\%, 12.5\%,25\%, 50\%$ and $100\%$ of the training set. In every training set partition we selected an equal number of tuples per number of sources. Additionally, the data set with $1.25\%$ is a subset of the data set with $2.5\%$ which is a subset of the data set with $5\%$ and hence forth as depicted in Fig. \ref{fig:DSDiag}. After training each model, we evaluated the performance on the test set per number of sources (see Fig. \ref{fig:DSDiag}). We tagged each model with a code which references the data set size used, the prefactor used, whether it was trained with MAE, MSE, Huber or inverse Huber and, in some case, we reference in the naming code the hyperparameters used. For instance, model 40E1 was trained with $1/40 = 2.5\%$ of the data set, an exponential prefactor and MAE while 80EIH was trained using $1/80=1.25\%$ of the data set, exponential prefactor and inverse Huber.

\begin{table}[hbtp]
\centering
{\renewcommand{\arraystretch}{1.3}
\caption{Models trained with $1.25\%$ of the training data set. During training of model 80S1\_a6000\_b1, we used a learning rate equal to $0.00002$. Models 80S1\_a4000\_b10\_sqrt, 80S1\_a6000\_b1\_sqrt and 80SH\_sqrt were trained by applying square root to the training data.}
\begin{tabular}{|c c c c c c c c|} 
 \hline
 Codename & w & a & b & $\delta$ & Loss Function & Epochs & App Trans \\ [0.5ex] 
 \hline\hline
 80E1 & 1 & - & - & - & Exp-MAE & 100 & - \\
 \hline
 80E2 & 1 & - & - & - & Exp-MSE & 100 & - \\
 \hline
 80EIH & 1 & - & - & - & Exp-IH & 100 & - \\
 \hline
 80S1\_a4000\_b10\_sqrt & - & 4000 & 10 & - & Step-MAE & 100 & sqrt \\
 \hline
 80S1\_a6000\_b1\_sqrt & - & 6000 & 1 & - & Step-MAE & 100 & sqrt \\
 \hline
 80S1\_a4000\_b10 & - & 4000 & 10 & - & Step-MAE & 100 & - \\
 \hline
 80S1\_a6000\_b1 & - & 6000 & 1 & - & Step-MAE & 100 & - \\
 \hline
 80S1\_a8000\_b1 & - & 8000 & 1 & - & Step-MAE & 100 & - \\
 \hline
 80S1\_a6000\_b10 & - & 6000 & 10 & - & Step-MAE & 100 & - \\
 \hline
 80S2 & - & 6000 & 10 & - & Step-MSE & 100 & - \\
 \hline
 80SH & - & 6000 & 1 & - & Step-Huber & 100 & - \\
 \hline
 80SH\_sqrt & - & 4000 & 10 & - & Step-Huber & 100 & sqrt \\
 \hline
 \end{tabular}
\label{table:1-2}
}
\end{table}

\begin{figure}[hbtp]
\centering
\includegraphics[width=2.10in]{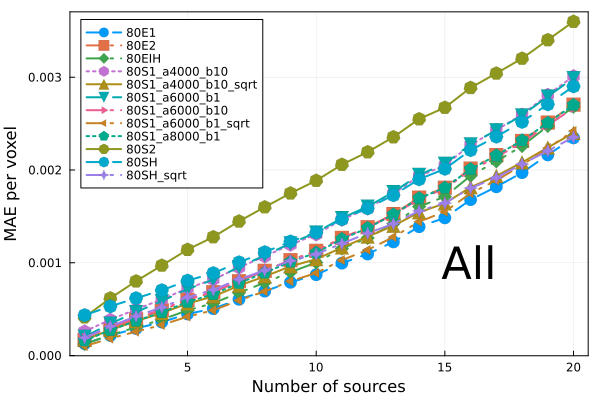}
\includegraphics[width=2.10in]{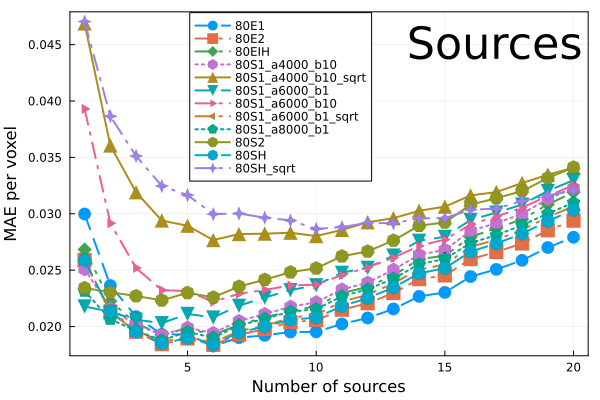}
\includegraphics[width=2.10in]{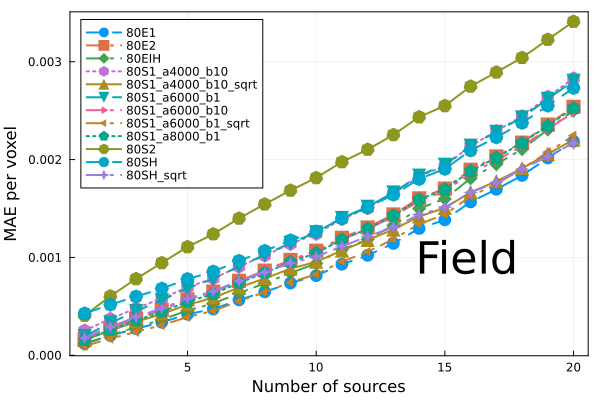}
\includegraphics[width=2.10in]{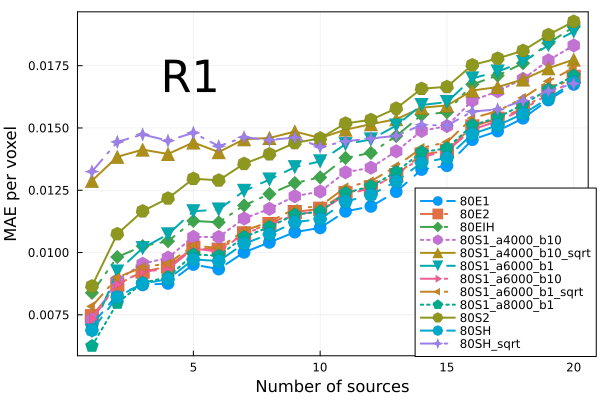}
\includegraphics[width=2.10in]{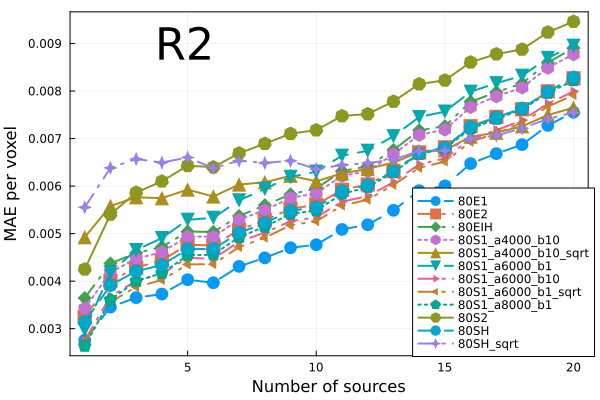}
\includegraphics[width=2.10in]{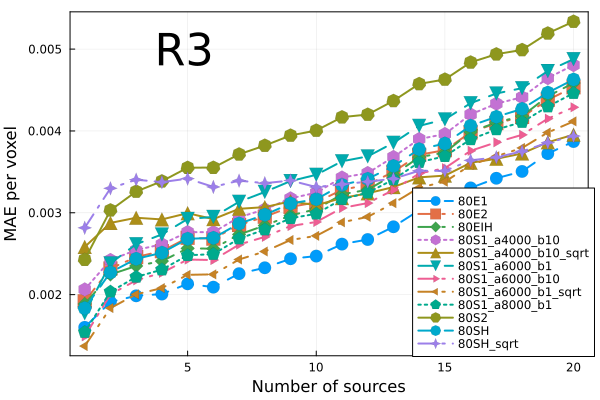}
\caption{ MAE \textit{vs} number of sources. Each curve correspond to different models (see legend) trained with $1.25\%$ of the data. Each plot correspond to a specific region in the lattice (see Fig. \ref{fig:inout0})} \label{fig:size_p1}
\end{figure}

In Fig. \ref{fig:size_p1} we show the MAE for models trained with $1.25\%$ of the data set. The hyperparameter selected for each model are displayed in Table \ref{table:1-2}. There are a number of features to highlight from Fig. \ref{fig:size_p1}. First, the largest absolute error occurs at the sources and decreases as one moves away from the sources. In addition, the error increases monotonically with the number of sources. Interestingly, as the number of sources increases, the error at the sources decreases rapidly and then increases slow and steadily. Most models performed quite similar, but overall model 80E1 performed best. However, the difference in performance is not significant as we will discuss next.

\begin{figure}[hbtp]
\centering
\includegraphics[width=2.10in]{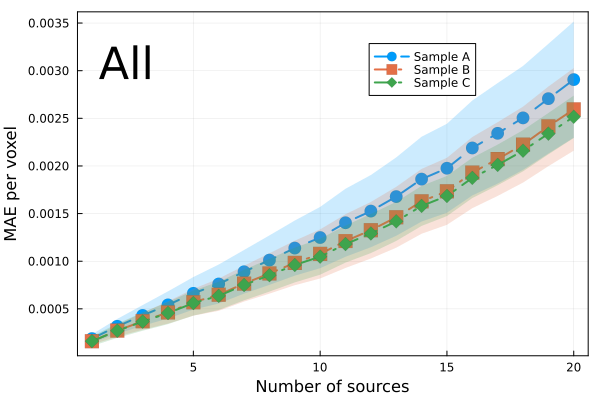}
\includegraphics[width=2.10in]{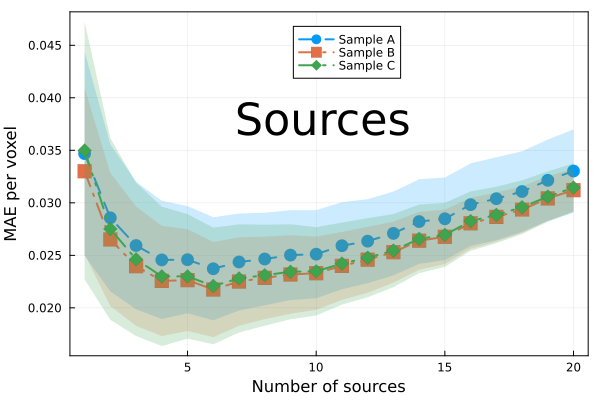}
\includegraphics[width=2.10in]{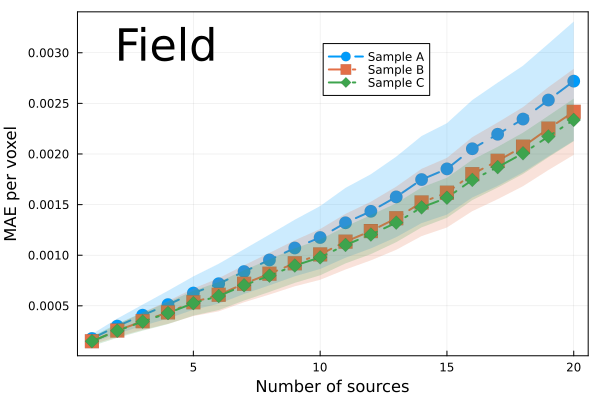}
\includegraphics[width=2.10in]{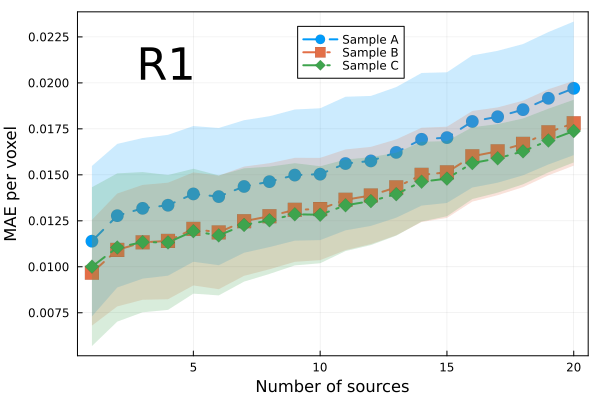}
\includegraphics[width=2.10in]{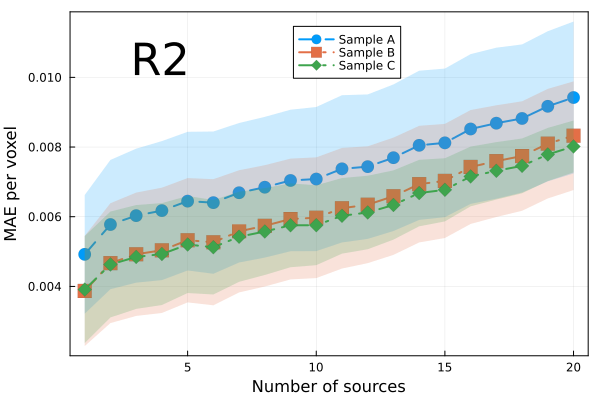}
\includegraphics[width=2.10in]{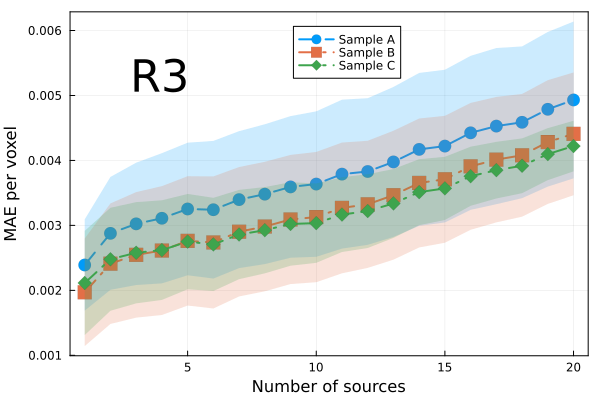}
\caption{ MAE \textit{vs} number of sources. Each curve correspond to the average ($\pm$ standard deviation) of a sample (generically tagged as A, B or C) composed by five models trained with the same $1.25\%$ data set but different initialization parameters and seeds. Each plot correspond to a specific region in the lattice (see Fig. \ref{fig:inout0})} \label{fig:size_p1_av}
\end{figure}

To assess the amplitude of fluctuations in performance due to stochasticity from initialization and training of the models, we trained three sets of five models each. All models were trained using the same hyperparameters. Each set was trained with a 1.25\% data set, such that there was no overlap between the data sets. In Fig. \ref{fig:size_p1_av} we show the average performance and standard deviation (in shadow) per set. Notice that there is no significant improvement between sets. Furthermore, the fluctuations per set are large enough to conclude that the different performances between the different models in Fig. \ref{fig:size_p1} can be attributed to fluctuations.


\begin{figure}[hbtp]
\centering
\includegraphics[width=2.10in]{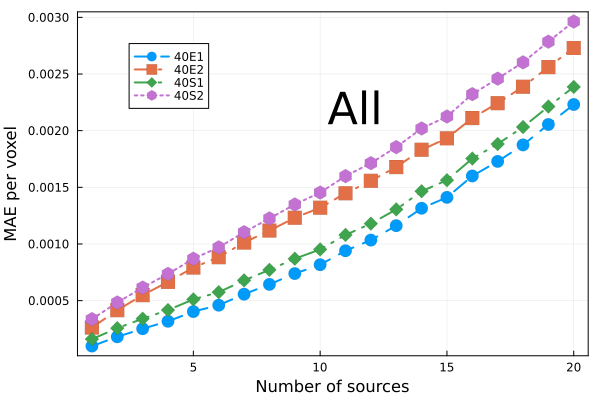}
\includegraphics[width=2.10in]{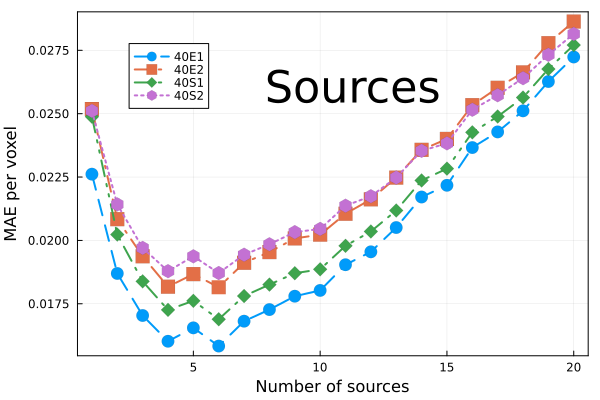}
\includegraphics[width=2.10in]{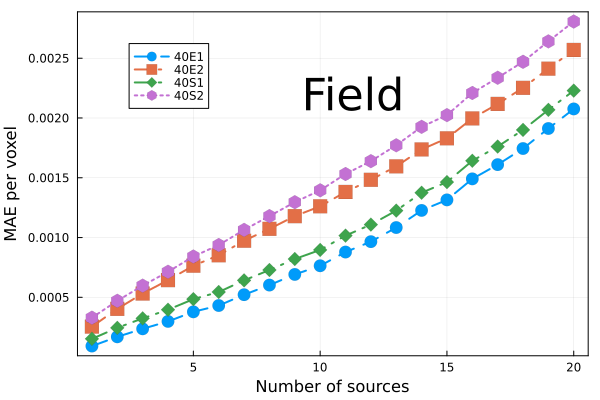}
\includegraphics[width=2.10in]{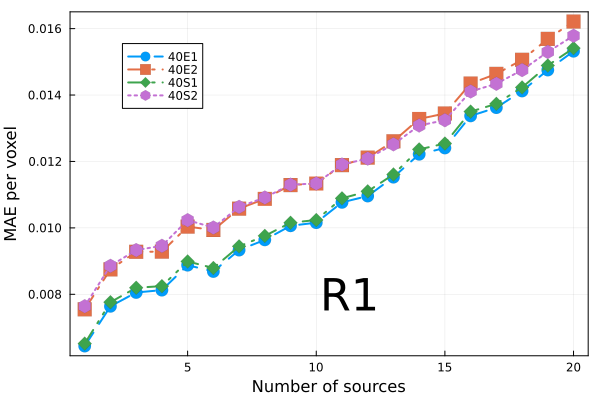}
\includegraphics[width=2.10in]{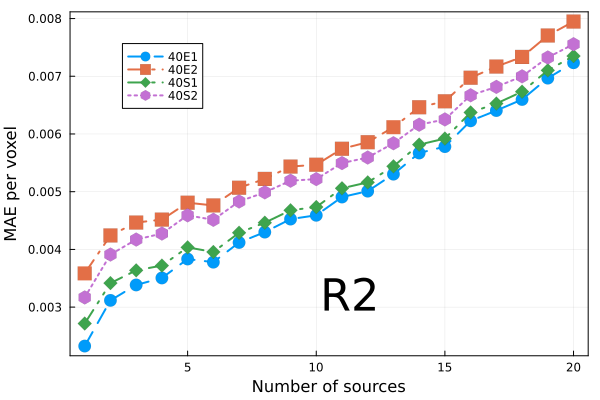}
\includegraphics[width=2.10in]{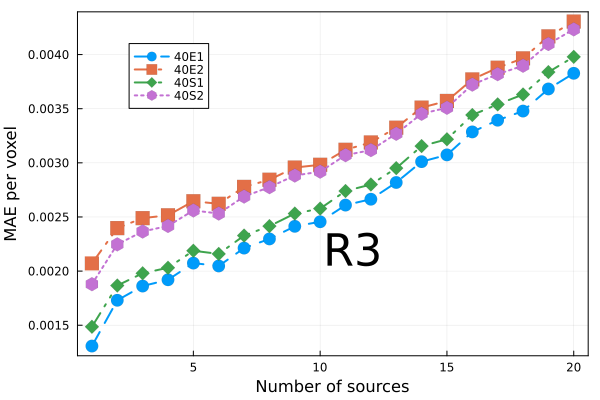}
\caption{ MAE \textit{vs} number of sources. Each curve correspond to different models trained with $2.5\%$ of the data. Each plot correspond to specific regions in the lattice (see Fig. \ref{fig:inout0})} \label{fig:size_p2}
\end{figure}

In Fig. \ref{fig:size_p2} we show the MAE for models trained with $2.5\%$ of the data set. The hyperparameter selected for each model are displayed in Table \ref{table:2-5}. Model 40E1, which was trained using the exponential weight MAE loss function, performed best in all the lattice regions and for all number of sources. However, notice that the model with the second best performance, model 40S1, was trained using the step-weighed MAE, while what the two worst two models have in common is that both were trained using MSE. Furthermore, the largest difference in performance between the best and the worst model, which happens in the sources, relative to the error is $\approx 0.2$. This suggests the different performance in models can be attributed to fluctuations and not necessarily the effect of the loss function. On the other hand, it was found that, in general, the performance decreases as the number of sources increases. As noted previously, in the case of the sources the performance first decreases, reaching a minimum around $5$ sources and it then increases with the number of sources. The appearance of a minimum in performance is a feature that we will come across throughout the present paper and we leave the discussion for later.

\begin{table}[hbtp]
\centering
{\renewcommand{\arraystretch}{1.3} 
\caption{Models trained with $2.5\%$ of the training data set.}
\begin{tabular}{|c c c c c c c c|} 
 \hline
 ID & w & a & b & $\delta$ & Loss Function & Epochs & App Trans \\ [0.5ex] 
 \hline\hline
 40E1 & 1 & - & - & - & Exp-MAE & 100 & - \\
 \hline
 40E2 & 1 & - & - & - & Exp-MSE & 100 & - \\
 \hline
 40S1 & - & 4000 & 10 & - & Step-MAE & 100 & - \\
 \hline
 40S2 & - & 4000 & 10 & - & Step-MSE & 100 & - \\
 \hline
 \end{tabular}
\label{table:2-5}
}
\end{table}


\begin{table}[hbtp]
\centering
{\renewcommand{\arraystretch}{1.3} 
\caption{Models trained with $5.0\%$ of the training data set.}
\begin{tabular}{|c c c c c c c c|} 
 \hline
 ID & w & a & b & $\delta$ & Loss Function & Epochs & App Trans \\ [0.5ex] 
 \hline\hline
 20E1 & 1 & - & - & - & Exp-MAE & 100 & - \\
 \hline
 20E2 & 1 & - & - & - & Exp-MSE & 100 & - \\
 \hline
 20EIH\_$\delta$0.5 & 1 & - & - & 0.5 & Exp-Inv-Huber & 100 & - \\
 \hline
 20EIH\_$\delta$0.05 & 1 & - & - & 0.05 & Exp-Inv-Huber & 100 & - \\
 \hline
 20EIH\_$\delta$0.15 & 1 & - & - & 0.15 & Exp-Inv-Huber & 100 & - \\
 \hline
 20S1 & - & 4000 & 10 & - & Step-MAE & 100 & - \\
 \hline
 20S2 & - & 4000 & 10 & - & Step-MSE & 100 & - \\
 \hline
 \end{tabular}
\label{table:5}
}
\end{table}


\begin{figure}[hbtp]
\centering
\includegraphics[width=2.10in]{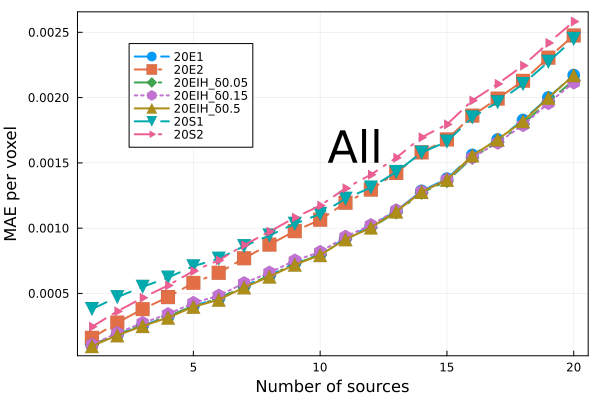}
\includegraphics[width=2.10in]{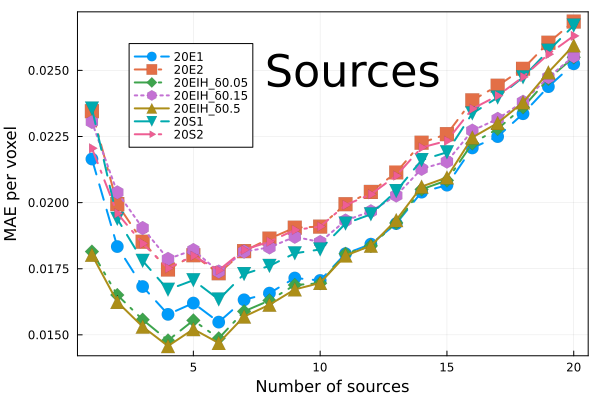}
\includegraphics[width=2.10in]{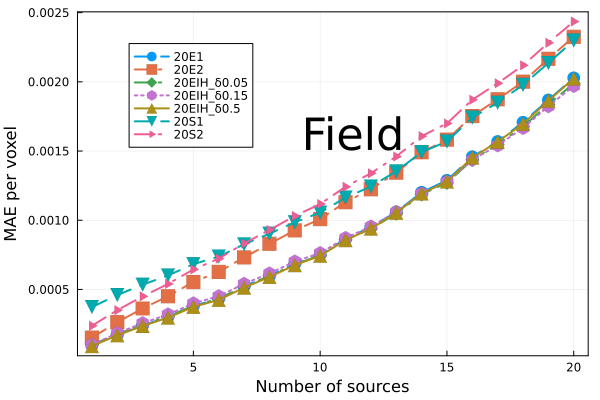}
\includegraphics[width=2.10in]{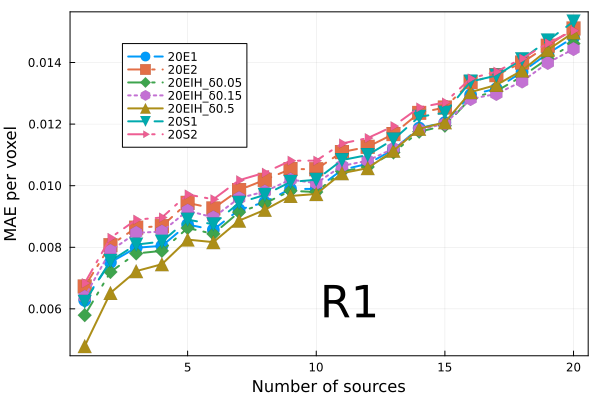}
\includegraphics[width=2.10in]{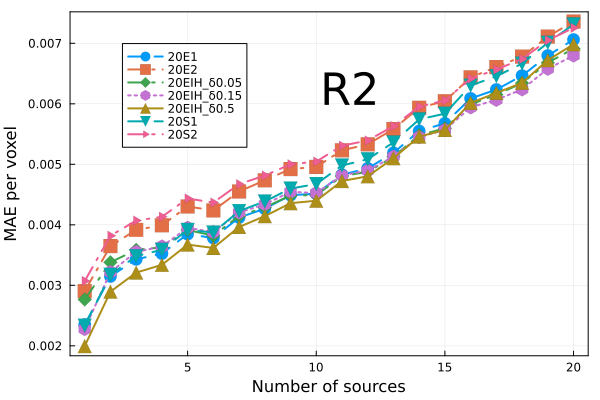}
\includegraphics[width=2.10in]{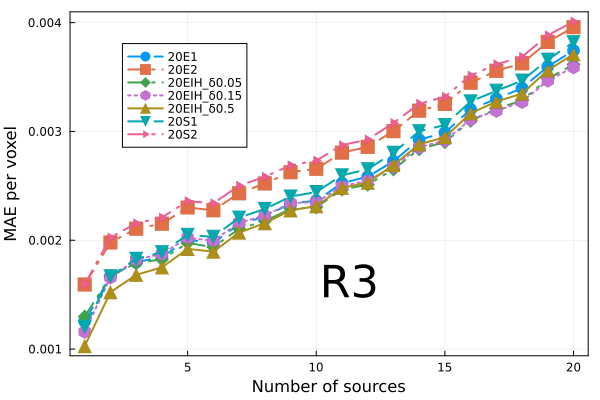}
\caption{ MAE \textit{vs} number of sources. Each curve correspond to different models trained with $5\%$ of the data. Each plot correspond to specific regions in the lattice (see Fig. \ref{fig:inout0})} \label{fig:size_p5}
\end{figure}

\begin{figure}[hbtp]
\centering
\includegraphics[width=2.10in]{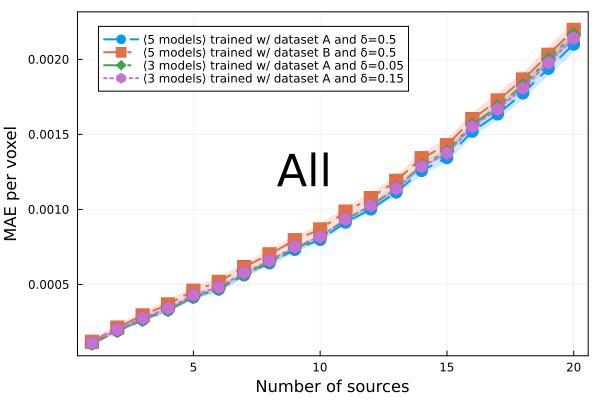}
\includegraphics[width=2.10in]{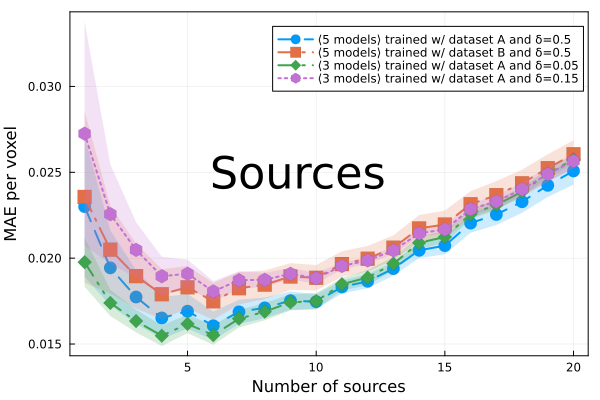}
\includegraphics[width=2.10in]{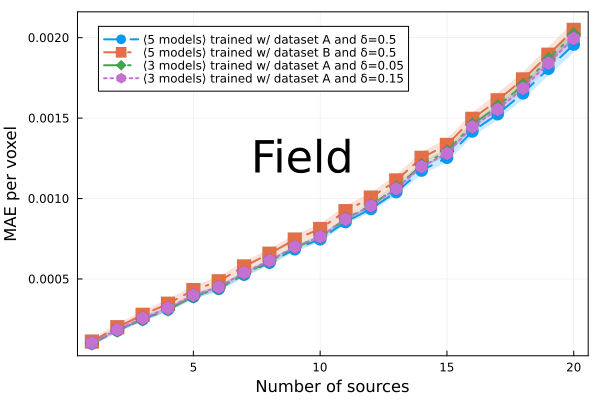}
\includegraphics[width=2.10in]{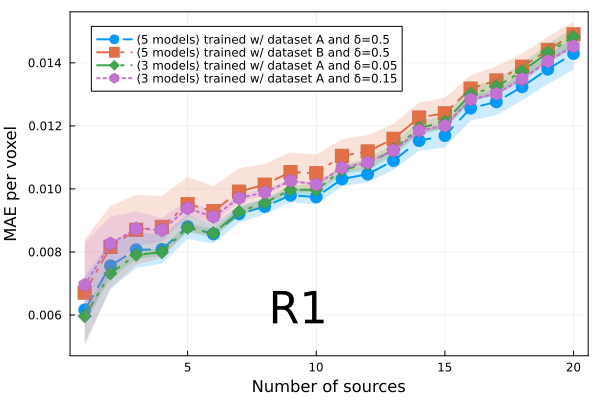}
\includegraphics[width=2.10in]{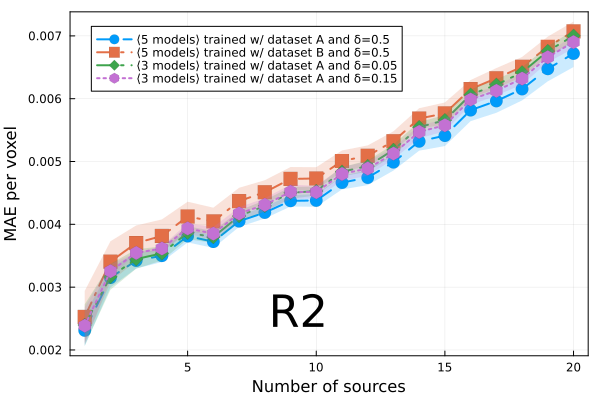}
\includegraphics[width=2.10in]{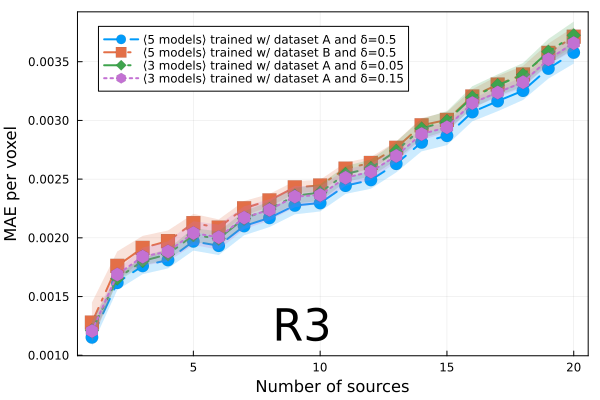}
\caption{ MAE \textit{vs} number of sources. Each curve correspond to the average ($\pm$ standard deviation) of a sample composed by three or five models (see legend) trained with $5.0\%$ size training set (generically called A or B, such that $A \cap B = \emptyset$). We used the inverse Huber loss function in all 16 models with $\delta$ fixed to $\lbrace 0.05, 0.15, 0.5 \rbrace$. The remaining hyperparameters were the same per curve but different initialization parameters and seeds. Each plot correspond to a specific region in the lattice (see Fig. \ref{fig:inout0}). } \label{fig:size_p5_av}
\end{figure}


\begin{table}[hbtp]
\centering
{\renewcommand{\arraystretch}{1.3} 
\caption{Models trained with $12.5\%$ of the training data set.}
\begin{tabular}{|c c c c c c c c|} 
 \hline
 ID & w & a & b & $\delta$ & Loss Function & Epochs & App Trans \\ [0.5ex] 
 \hline\hline
 8E1 & 1 & - & - & - & Exp-MAE & 100 & - \\
 \hline
 8E2 & 1 & - & - & - & Exp-MSE & 100 & - \\
 \hline
 8S1 & - & 4000 & 10 & - & Step-MAE & 100 & - \\
 \hline
 8S2 & - & 4000 & 10 & - & Step-MSE & 100 & - \\
 \hline
 \end{tabular}
\label{table:12}
}
\end{table}


\begin{figure}[hbtp]
\centering
\includegraphics[width=2.10in]{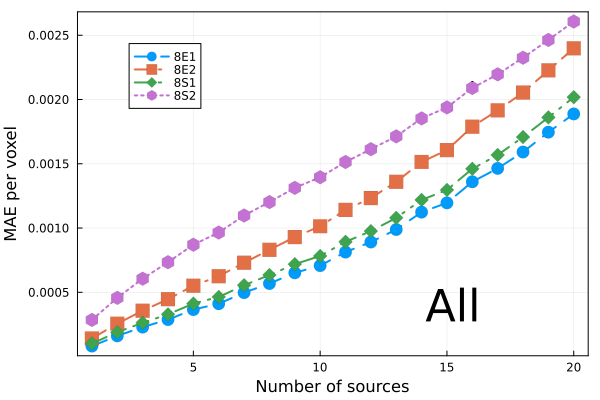}
\includegraphics[width=2.10in]{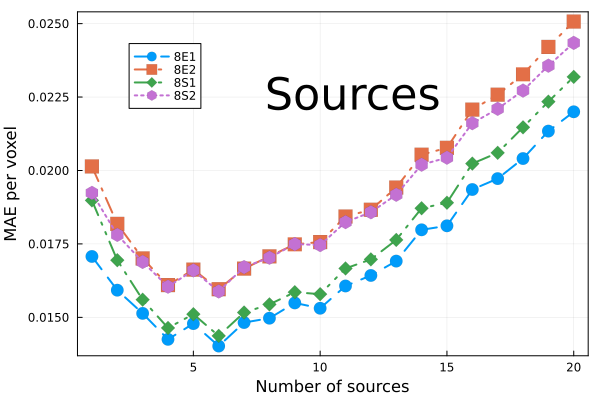}
\includegraphics[width=2.10in]{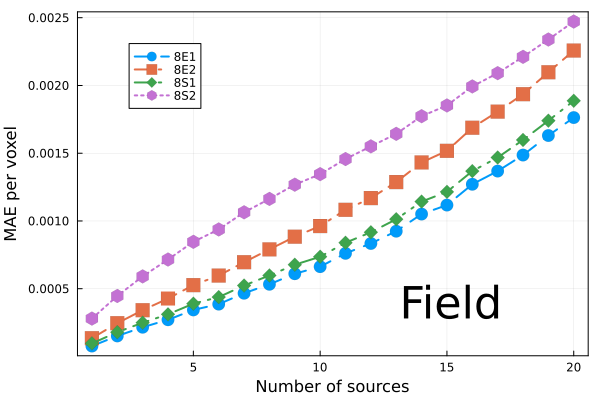}
\includegraphics[width=2.10in]{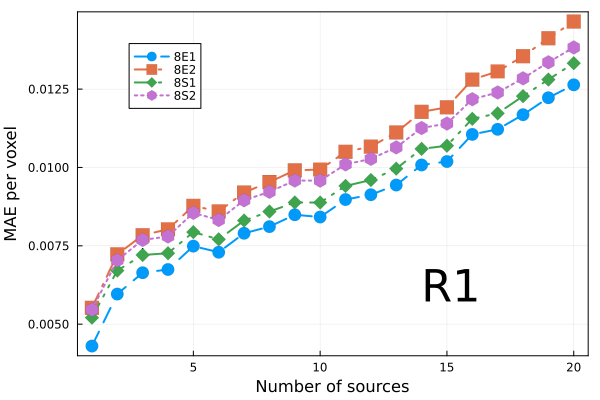}
\includegraphics[width=2.10in]{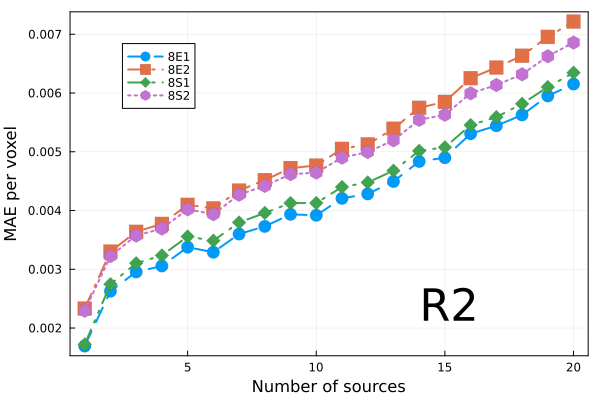}
\includegraphics[width=2.10in]{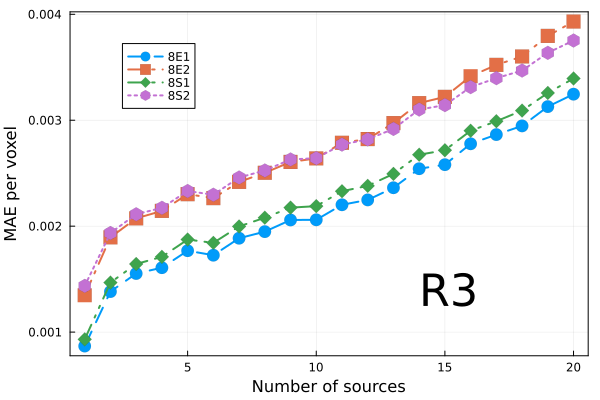}
\caption{ MAE \textit{vs} number of sources. Each curve correspond to different models (see legend) trained with $12.5\%$ of the data. Each plot correspond to specific regions in the lattice (see Fig.  \ref{fig:inout0}))} \label{fig:size_p12}
\end{figure}


\begin{table}[hbtp]
\centering
{\renewcommand{\arraystretch}{1.3} 
\caption{Models trained with $25\%$ of the training data set. Model 4E2-2 was trained adding Gaussian noise w/ standard deviation equal to $0.01$ on the groundtruth. In training model 4R, each epoch selected either step or exp prefactor randomly with probability $0.8$ for the former and $0.2$ for the latter. In training model 4T-1, the loss function toggled between step and exp prefactor, starting with exp. In training model 4T-5, the loss function toggled between step and exp prefactor every 5 epochs. }
\begin{tabular}{|c c c c c c c c|} 
 \hline
 ID & w & a & b & $\delta$ & Loss Function & Epochs & App Trans \\ [0.5ex] 
 \hline\hline
 4E1 & 1 & - & - & - & Exp-MAE & 100 & - \\
 \hline
 4E2 & 1 & - & - & - & Exp-MSE & 89 & - \\
 \hline
 4R & 1 & 4000 & 10 & - & MAE & 84 & - \\
 \hline
 4S1 & - & 4000 & 10 & - & Step-MAE & 100 & - \\
 \hline
 4SH & - & 4000 & 10 & - & Step-Huber & 100 & - \\
 \hline
 4T-1 & - & 4000 & 10 & - & Step-MSE & 100 & - \\
 \hline
 4T-5 & - & 4000 & 10 & - & Step-MSE & 100 & - \\
 \hline
 \end{tabular}
\label{table:25}
}
\end{table}

\begin{figure}[hbtp]
\centering
\includegraphics[width=2.10in]{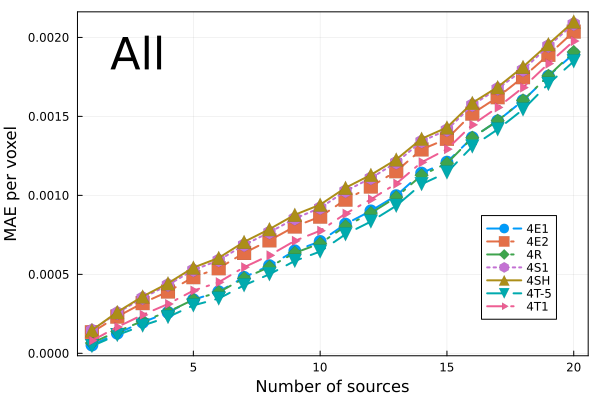}
\includegraphics[width=2.10in]{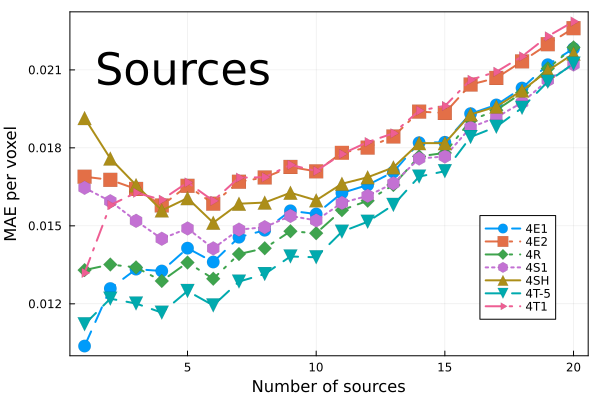}
\includegraphics[width=2.10in]{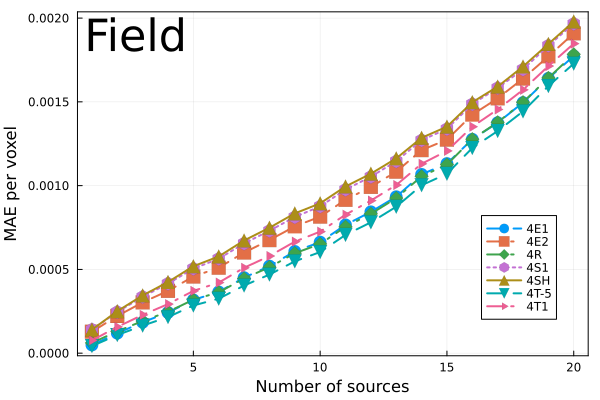}
\includegraphics[width=2.10in]{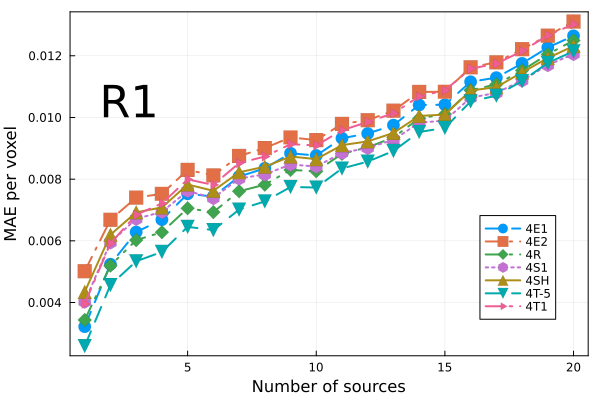}
\includegraphics[width=2.10in]{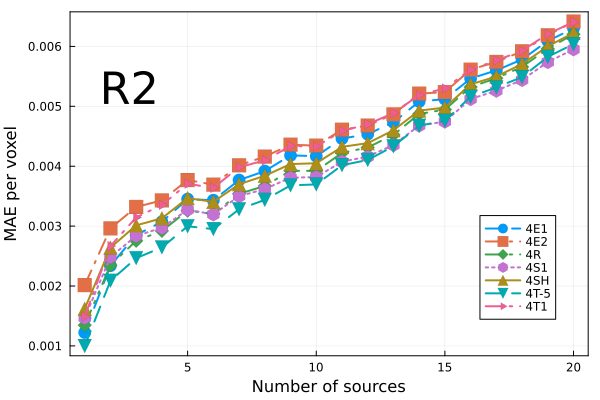}
\includegraphics[width=2.10in]{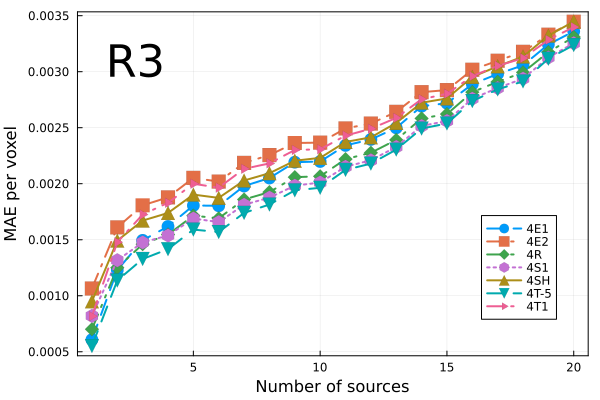}
\caption{ MAE \textit{vs} number of sources. Each curve correspond to different models (see legend) trained with $25\%$ of the data. Each plot correspond to specific regions in the lattice (see Fig.  \ref{fig:inout0}))} \label{fig:size_p25}
\end{figure}


\begin{table}[hbtp]
\centering
{\renewcommand{\arraystretch}{1.3} 
\caption{Models trained with $50\%$ of the training data set.}
\begin{tabular}{|c c c c c c c c|} 
 \hline
 ID & w & a & b & $\delta$ & Loss Function & Epochs & App Trans \\ [0.5ex] 
 \hline\hline
 2E1 & 1 & - & - & - & Exp-MAE & 64 & - \\
 \hline
 2S1 & - & 4000 & 10 & - & Step-MAE & 81 & - \\
 \hline
 \end{tabular}
\label{table:50}
}
\end{table}


\begin{figure}[hbtp]
\centering
\includegraphics[width=2.10in]{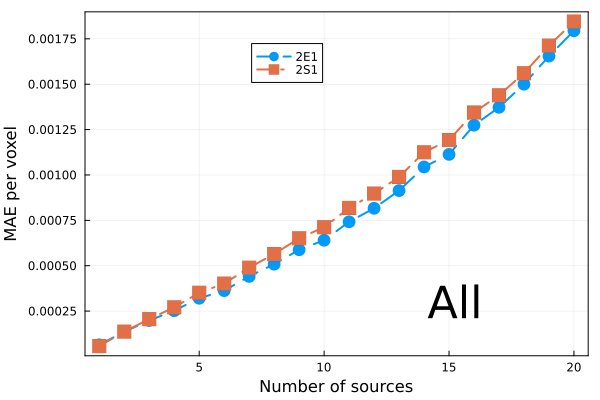}
\includegraphics[width=2.10in]{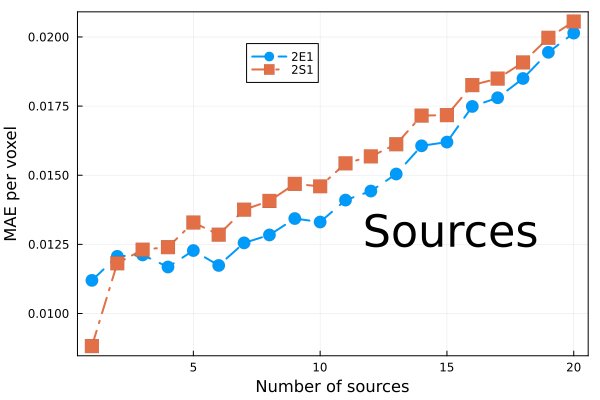}
\includegraphics[width=2.10in]{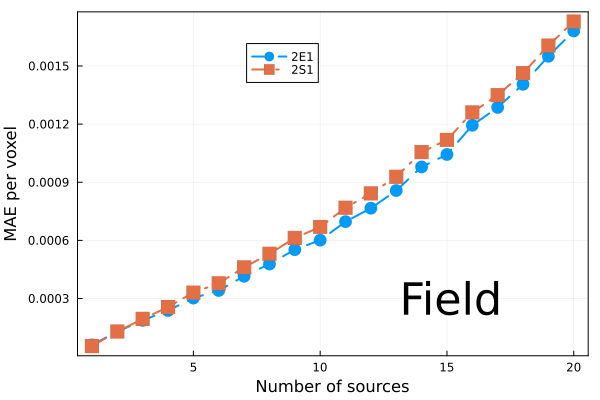}
\includegraphics[width=2.10in]{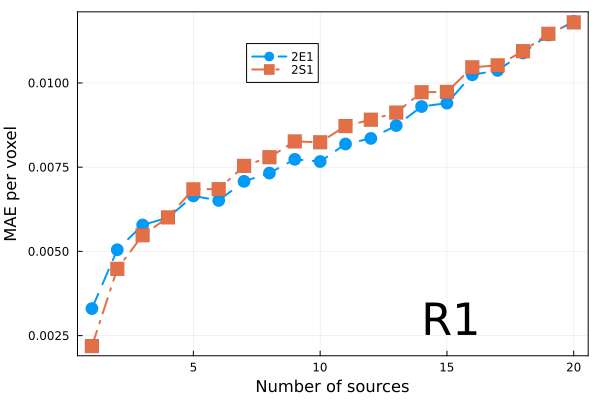}
\includegraphics[width=2.10in]{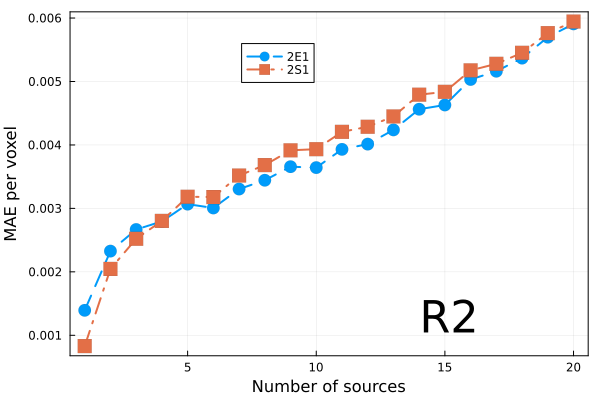}
\includegraphics[width=2.10in]{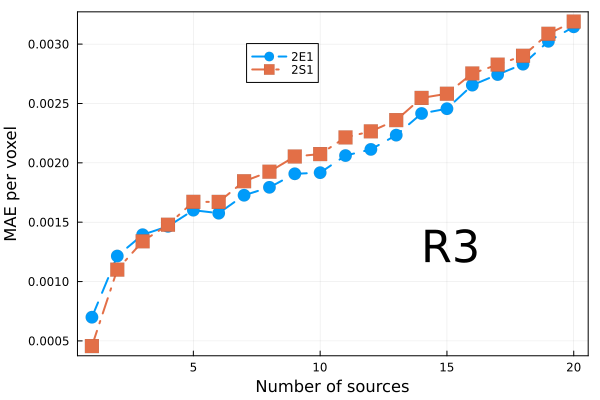}
\caption{ MAE \textit{vs} number of sources. Each curve correspond to different models (see legend) trained with $50\%$ of the data. Each plot correspond to specific regions in the lattice (see Fig.  \ref{fig:inout0}))}  \label{fig:size_p50}
\end{figure}


\begin{table}[hbtp]
\centering
{\renewcommand{\arraystretch}{1.3} 
\caption{Models trained with $100\%$ of the training data set.}
\begin{tabular}{|c c c c c c c c|} 
 \hline
 ID & w & a & b & $\delta$ & Loss Function & Epochs & App Trans \\ [0.5ex] 
 \hline\hline
 1E1 & 1 & - & - & - & Exp-MAE & 56 & - \\
 \hline
 1S1 & - & 4000 & 10 & - & Step-MAE & 40 & - \\
 \hline
 \end{tabular}
\label{table:100}
}
\end{table}


\begin{figure}[hbtp]
\centering
\includegraphics[width=2.10in]{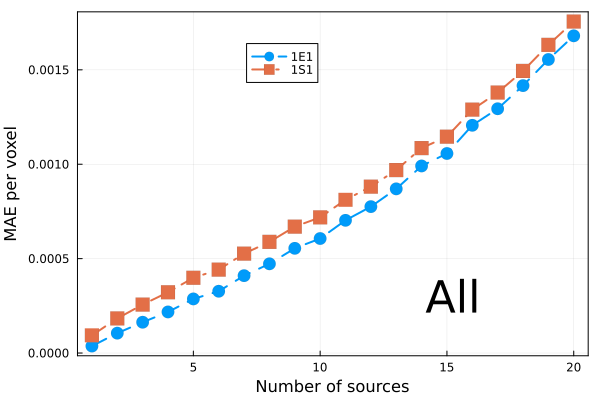}
\includegraphics[width=2.10in]{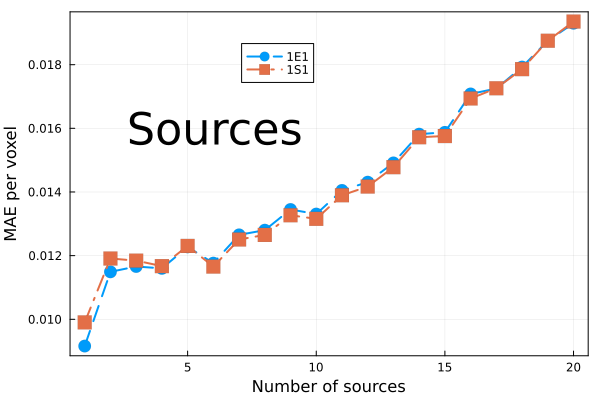}
\includegraphics[width=2.10in]{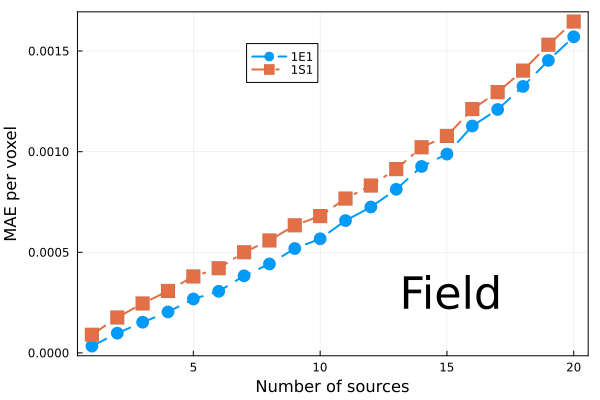}
\includegraphics[width=2.10in]{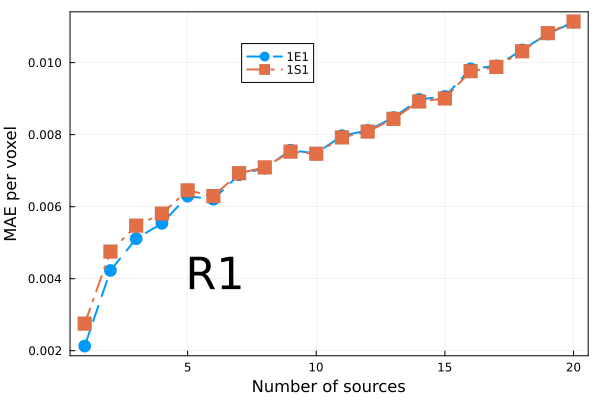}
\includegraphics[width=2.10in]{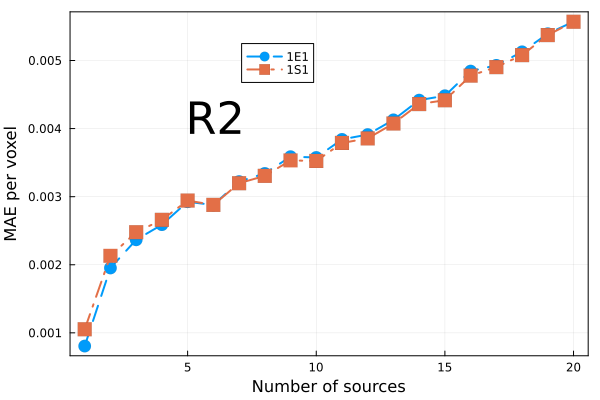}
\includegraphics[width=2.10in]{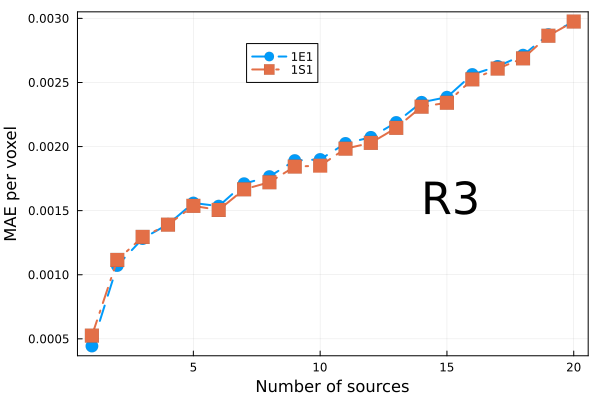}
\caption{ MAE \textit{vs} number of sources. Each curve correspond to different models (see legend) trained with $100\%$ of the data. Each plot correspond to specific regions in the lattice (see Fig.  \ref{fig:inout0}))} \label{fig:size_p100}
\end{figure}

In Fig. \ref{fig:size_p5} we show the MAE for models trained with $5\%$ of the data set. The hyperparameter selected for each model are displayed in Table \ref{table:5}. Notice that the same general trends discussed previously also holds in this case. In particular, models 20EIH-\# performed best, which were trained using the exponential weight inverse Huber loss function. On the other hand, model 20E2 had the worst performance. To assess the amplitude of fluctuations in performance due to stochasticity from initialization and training of the models, we trained two sets of five models each. All models were trained using the same hyperparameters. Each set was trained with a 5\% data set, such that there was no overlap between the training sets. We also trained two sets more composed by 3 models each, using different $\delta$-hyperparameter. In Fig. \ref{fig:size_p5_av} we show the average performance and standard deviation (in shadow) per set. Notice that there is no significant improvement between sets. However, the fluctuations per set have reduced quite substantially compared to the 1.25\% case shown in Fig. \ref{fig:size_p1_av} and discussed previously. In this regard, the different performances between the different models in Fig. \ref{fig:size_p5} can be attributed to some extent to fluctuations, however notice that model 20E2 can be taken as an outlier. 

We trained multiple models using 12\%, 25\%, 50\% and 100\% of the data set. The models hyperparameters are specified in Tables \ref{table:12}, \ref{table:25}, \ref{table:50} and \ref{table:100}, respectively. The performances are shown in Figs. \ref{fig:size_p12}, \ref{fig:size_p25}, \ref{fig:size_p50} and \ref{fig:size_p100}, respectively. A repeating feature is that the error increases with the number of sources in all cases, as discussed previously. Moreover, models trained using the exp prefactor typically outperform the rest. We also trained a model, 4R, that each epoch randomly selects between exp-prefactor and step-prefactor with probability $0.2$ and $0.8$, respectively. We also trained two models, 4T-1 and 4T-5, that toggle between step- and exp-prefactor every 1 and 5 epochs, respectively. Notice that models 4R and 4T-5 outperform the rest, including 4E1. The rationale behind models 4T- and 4R is that perturbing the model over the loss function landscape, via toggling or randomly choosing between the loss functions, can help the model find a lower local minimum. In the case of 50\% and 100\% data set size, we only trained two models per data set size as we wanted to compare performance between step- and exp-prefactor. The exponential weighing prefactor show better performance in all regions of the lattice, as shown in Figs. \ref{fig:size_p50} and \ref{fig:size_p100}.

\begin{figure}[hbtp]
\centering
\includegraphics[width=2.10in]{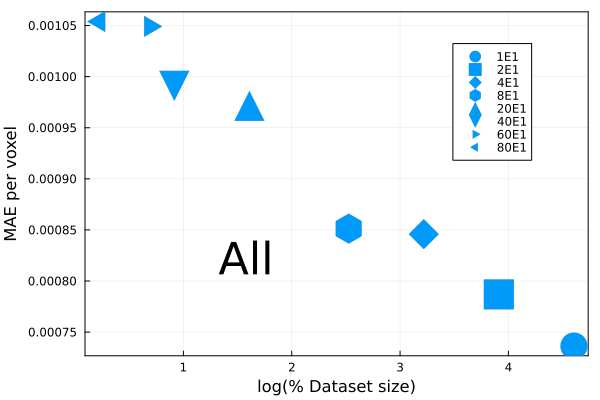}
\includegraphics[width=2.10in]{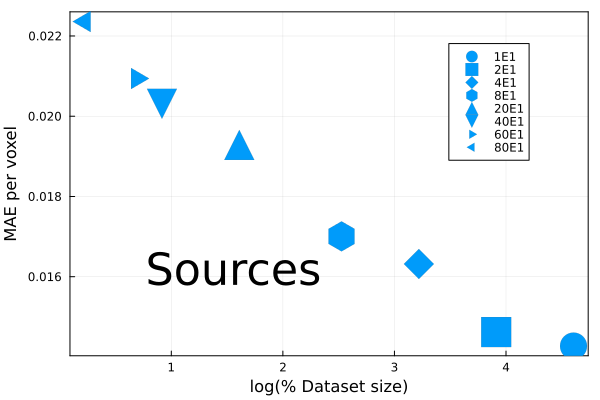}
\includegraphics[width=2.10in]{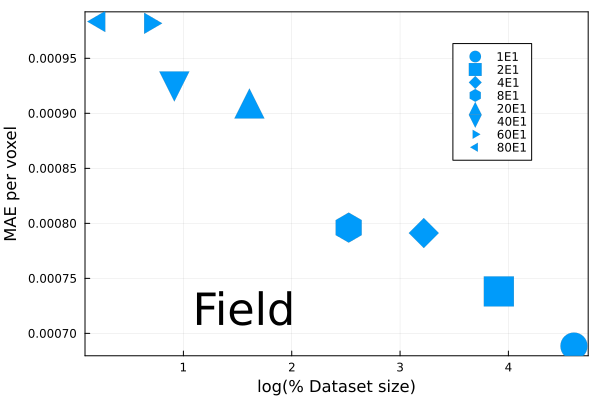}
\includegraphics[width=2.10in]{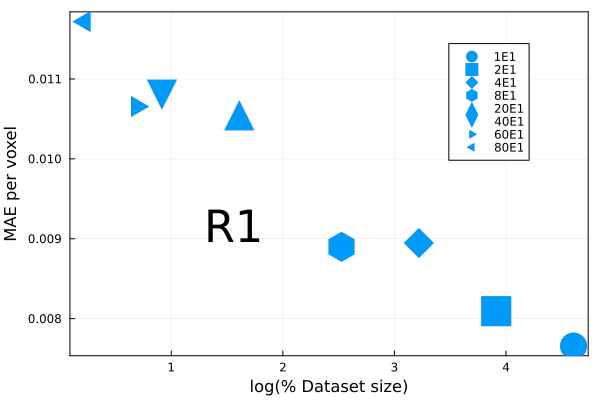}
\includegraphics[width=2.10in]{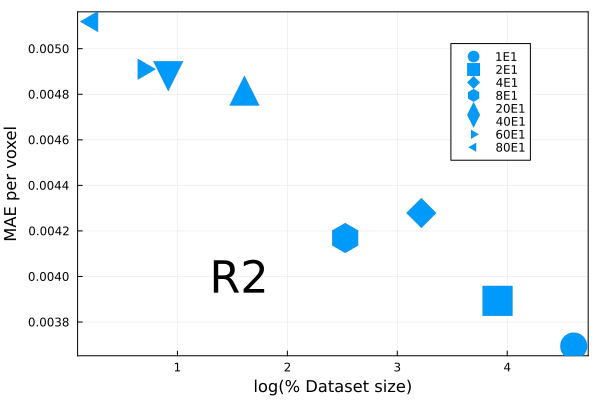}
\includegraphics[width=2.10in]{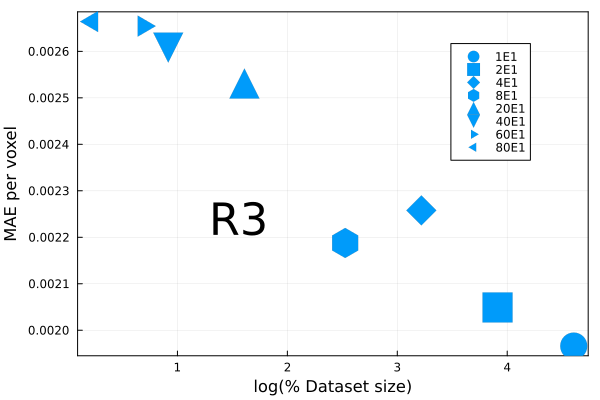}
\caption{ MAE averaged over number of sources \textit{vs} \% of training set size. Each point correspond to the models that performed best per amount of data (see legend). Each plot correspond to specific regions in the lattice (see Fig. \ref{fig:inout0}))} \label{fig:size_bmMean}
\end{figure}

In the previous the different performances for different data set size and different loss functions were shown and discussed. We showed that increasing the data set size decreases the performance fluctuations. We also showed that, in general, different loss functions yield similar performance. However, increasing the data set size improves the model performance. In Fig. \ref{fig:size_bmMean} we show the performance of models trained with different data set size and with the same loss function, namely, exponential-weighed MAE. Notice the performance has a logarithmic dependence on the training set size. The main takeaways are, first, models trained with exponential prefactor and MAE consistently show better performance; second, changing loss functions between epochs can have a positive effect in performance as seen by models 4R and 4T-5; third, as expected the error decreases with the training set size; fourth, as the training size increases the performance increases but the rate decreases. The later is surprising, because the high dimensionality of the problem means that even our largest training set significantly undersamples the space of possible configurations. This saturation points to a possible fundamental limit in the ability of the encoder-decoder network to solve the diffusion problem and suggests that other networks might be more effective. 



\subsection{Inference}
In this section we present and discuss the results for prediction by inference. While in the previous section we focused on performance for different data set size, here we study performance of models trained on a data set with a fixed number of sources. In this sense, we look at how well a model performs when given an input with an arbitrary number of sources different to the number of sources the model was trained on. This is particularly relevant since, by brute estimation, the configurational space is considerably large and exponentially dependent on the number of sources, however, by taking into account the configurational space symmetries one could construct the training set where the number of samples is depend on the number of sources. We found that there is an optimum, \textit{i.e.}, models trained on a fixed and different number of sources perform differently.

\begin{figure}[hbtp]
\centering
\includegraphics[width=2.10in]{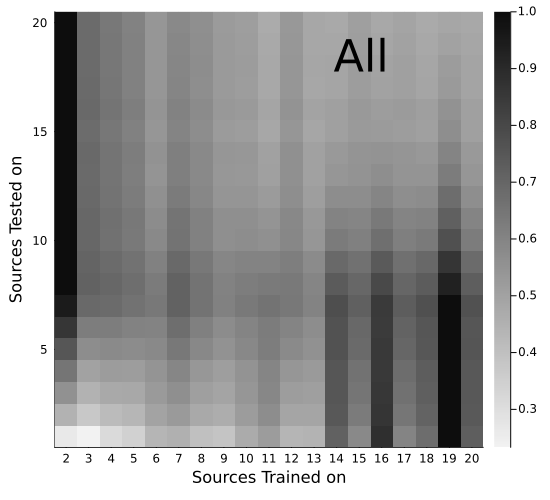}
\includegraphics[width=2.10in]{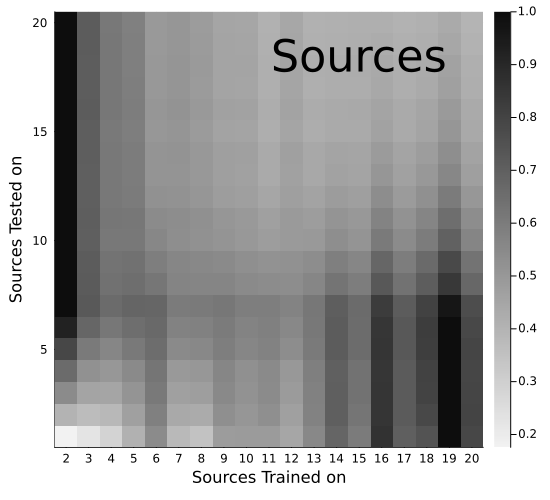}
\includegraphics[width=2.10in]{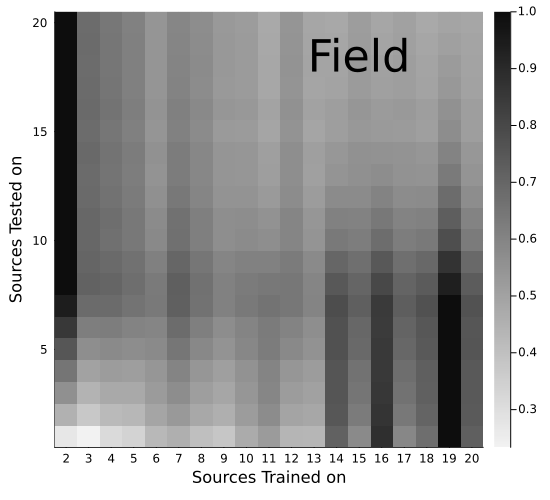}
\includegraphics[width=2.10in]{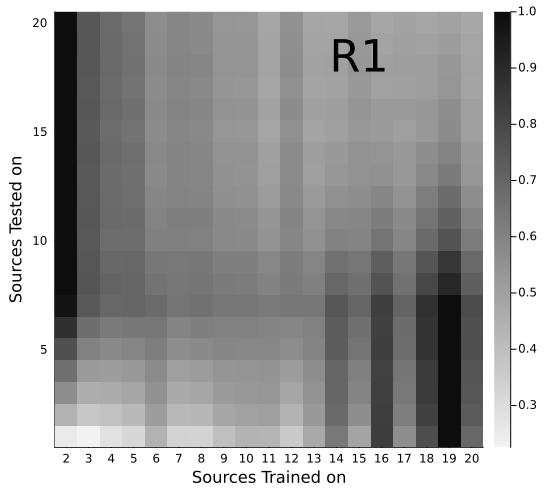}
\includegraphics[width=2.10in]{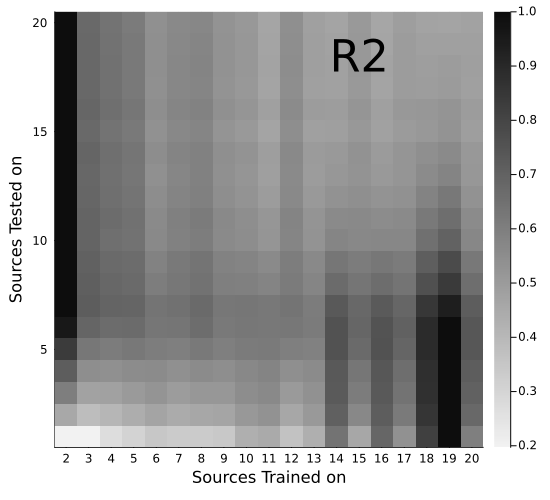}
\includegraphics[width=2.10in]{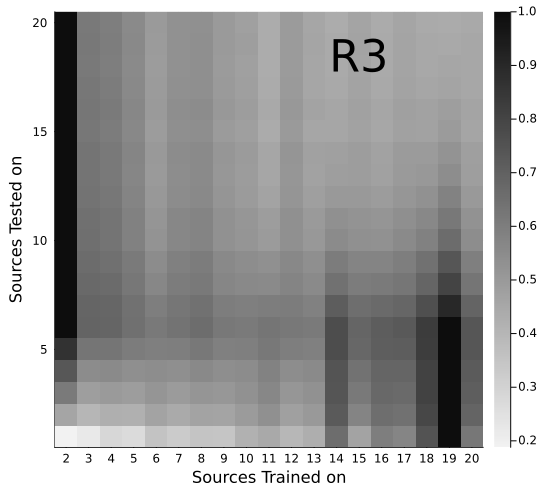}
\caption{ MAE of models trained on a fixed number of sources $P$ (X axis) and tested on a fixed number of sources $M$ (Y axis). Each row is normalized by dividing by the maximum value per row. Each row shows the MAE per model tested on a fixed number of sources. Each plot correspond to specific regions in the lattice (see Fig. \ref{fig:inout0}))} \label{fig:inference}
\end{figure}

We trained 20 models on each of the 20 data sets \textit{one-on-one}. Each model was trained for 100 epochs and using the exponentially weighed MAE as the loss function. To measure the inference prediction, we test the prediction of each model over each of the 20 test sets. Fig. \ref{fig:inference} shows a density plot of the normalized MAE for different regions in the lattice. The normalization is per row, \textit{i.e.}, for a fixed number of sources tested on (Y axis), each row yields the normalized MAE corresponding to each of the 19 models trained using a fixed number of sources (we have not included the model trained using 1 source as the MAE is an order of magnitude larger). These plots show that there is not a model that outperforms the rest for all test sets, rather models trained with $\sim 10-12$ perform better.

\begin{figure}[hbtp]
\centering
\includegraphics[width=1.10in]{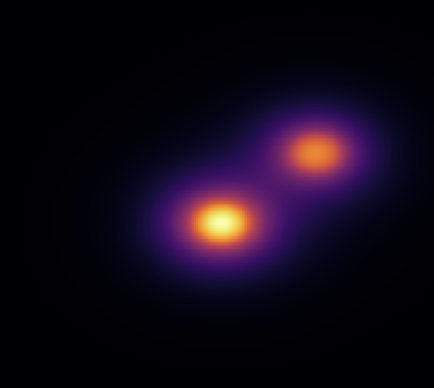}
\includegraphics[width=1.15in]{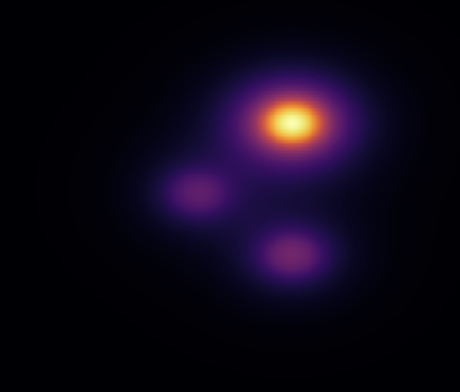}
\includegraphics[width=1.14in]{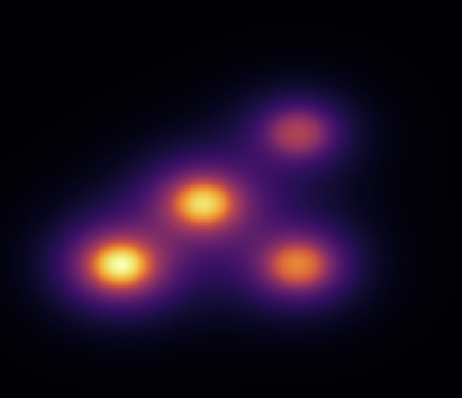}
\includegraphics[width=1.19in]{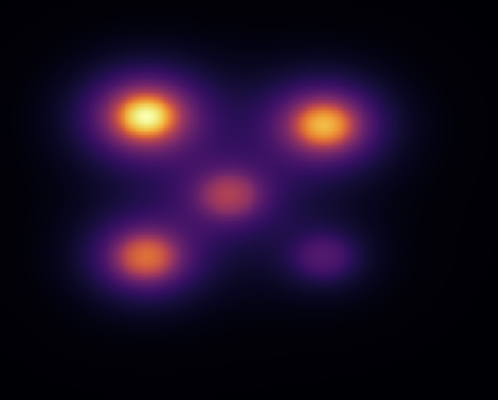}
\includegraphics[width=1.28in]{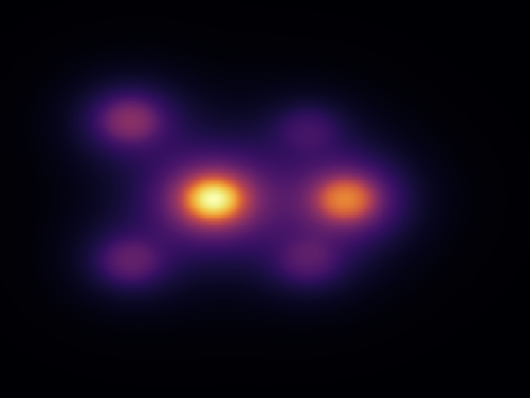}
\includegraphics[width=1.2in]{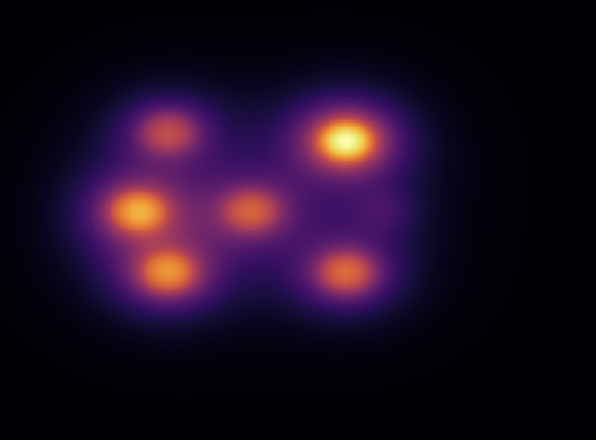}
\includegraphics[width=1.16in]{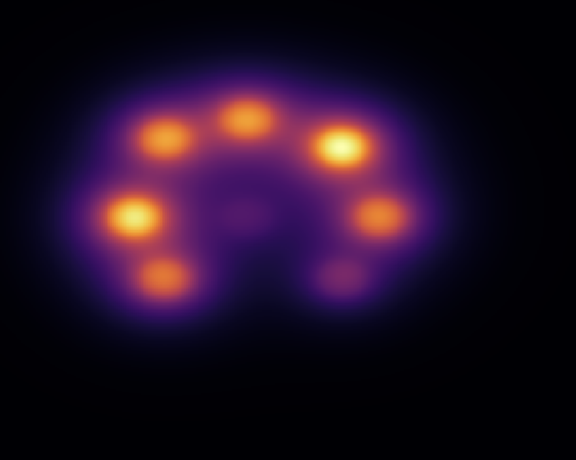}
\includegraphics[width=1.2in]{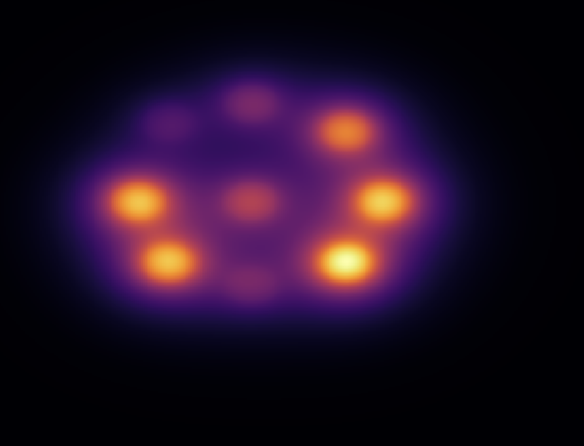}
\caption{ Two to nine sources clustered. Models trained using a data set with a small number of sources will not learn the to predict the steady-state of a large number of sources clustered. Conversely, models trained using a data set with a large number of sources will not learn the to predict the steady-state of a small number of sources clustered. Models trained using a data set capable of generating these configurations will yield better performance.} \label{fig:sourcesTogether}
\end{figure}

\begin{figure}[hbtp]
\centering
\includegraphics[width=2.10in]{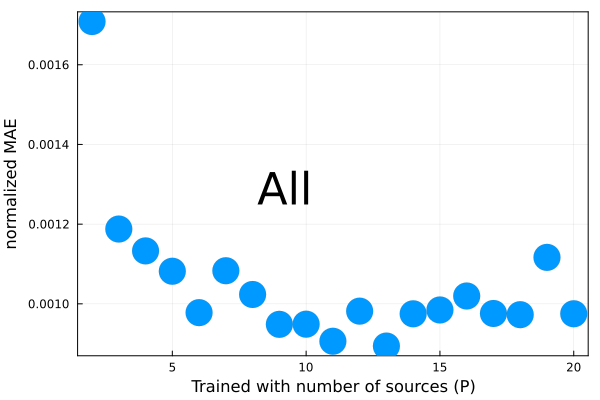}
\includegraphics[width=2.10in]{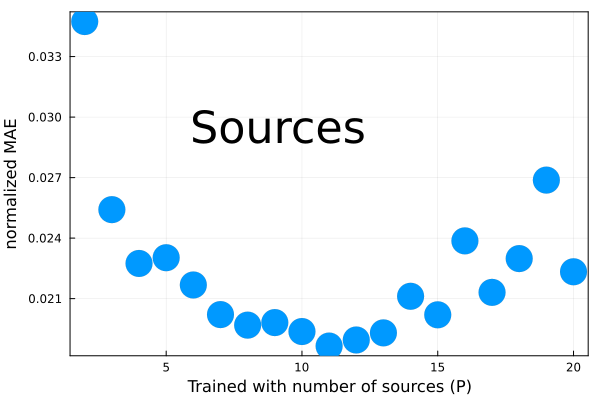}
\includegraphics[width=2.10in]{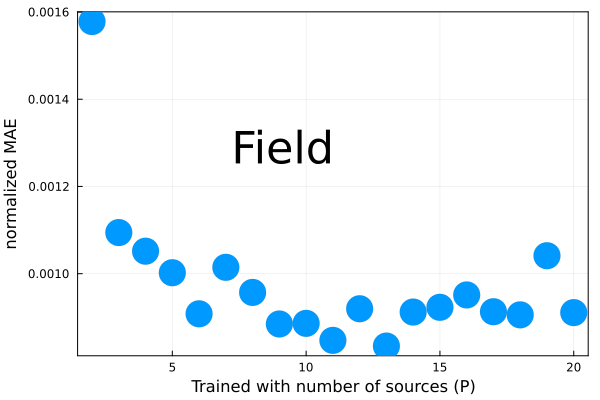}
\includegraphics[width=2.10in]{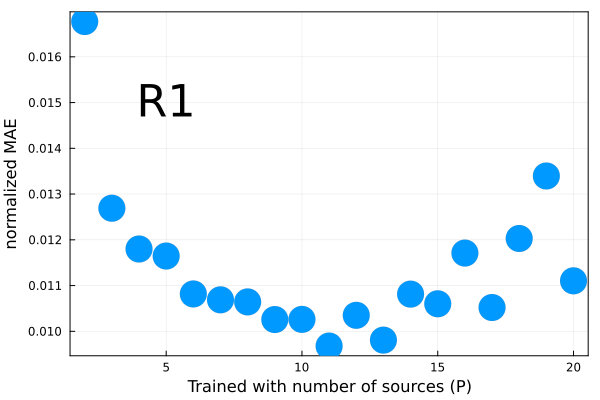}
\includegraphics[width=2.10in]{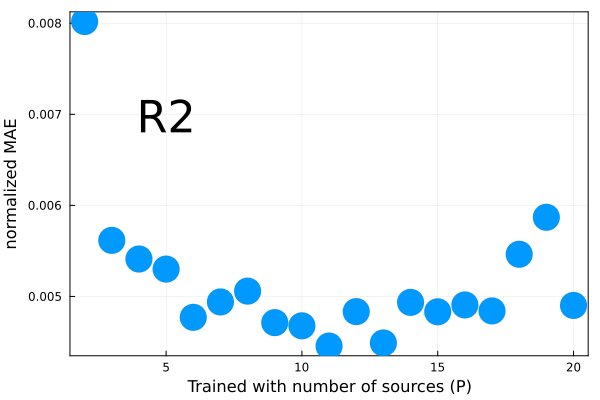}
\includegraphics[width=2.10in]{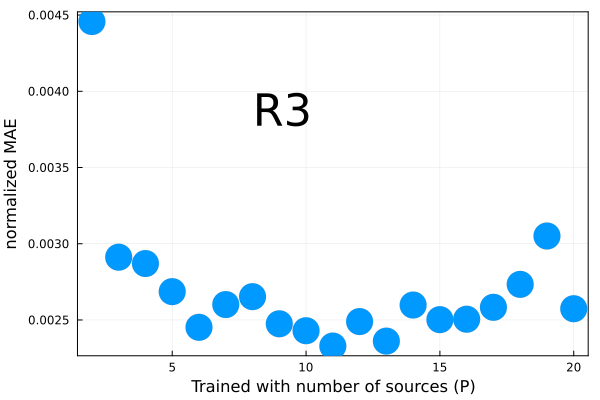}
\caption{Mean average error for a test set with containing samples with all number of sources for a network trained with a different fixed number of sources $P$. Each data point corresponds to a network trained with a fixed number of sources $P$ (x-axis).} \label{fig:inference2}
\end{figure}

We have averaged each of the non-normalized rows in Fig. \ref{fig:inference} and we have plotted the values in Fig. \ref{fig:inference2}. The model that yields the minimum error will depend on the region in the lattice we are considering, yet overall the optimum lies in models trained between 10 and 13 sources. The previous results suggest that the different configurations obtained from considering 10 to 13 sources in a lattice is sufficient to extrapolate to the different configuration that can arise from considering more sources. 
We can understand this in the following manner. Notice that what the model is learning is the field generated by the superposition of sources at different relative distances. In this regard, a model trained using a data set with a small number of sources will not have a large number of sources clustered and, hence, will never learn the field generated from a large number of sources. On the other hand, a data set with a large number of sources will most likely have a large number of clustered sources and lack small clusters of sources and, thus, a model trained using this data set will not learn the field generated from a few close sources. The data sets with number of sources around 12 are such that contain different clusters as depicted in Fig. \ref{fig:sourcesTogether}. The results in Figs. \ref{fig:inference} and \ref{fig:inference2} are in fact suggesting redundancy in the data set, \textit{i.e.}, a better curated data set would be comprised by different data set sizes dependent upon the number of sources. 


\section{Discussion}
Deep diffusion surrogates can aid in obtaining the steady-state solution for multiscale modeling. In this paper we looked at 20 sources randomly placed in 2D lattice. It's still far less complex than the real problem, \textit{e.g.}, simulation of a vascular system. Nevertheless, this is a step forward in that direction. We have shown that increasing the number of sources already pose a number of challenges in different aspects. We showed how the network architecture, the training set structure and size, the loss function, the hyperparameters for training algorithms and defining metrics to evaluate the task-specific performance of the trained network (which may differ from the loss function used in training) are all aspects that affect the final product. In the case of the NN architecture we argued that the encoder-decoder CNN architecture performs well not due to data compression, as believed by many, rather to data transformation akin to Fourier transform. However a more rigorous proof is, both required and desired. That's not to say that other architecture should not be able to perform well in predicting the steady-state solution. In fact, in prior work \cite{toledo2021deep} we combined an ED-CNN and a CNN for a similar task and found that the CNN improved the performance by reinforcing the sources in the prediction. The results shown in this paper highlight that the largest absolute error occurs at and near the sources. Therefore, we reckon the architecture such as a UNet \cite{ronneberger2015u} as a good candidate for this task and we leave it for future work.

We considered different loss functions and compared the different performances due to the different loss functions. The wide numeric range for input and output of neural networks makes the analysis very sensitive to choices in the loss function. Our results suggest that the loss function can have a significant effect on the model's performance. However, we showed that the data set size has a greater effect in the model's performance and, furthermore, the performance associated to the loss function depends on the data set size. We also showed how a large enough data set reduces the performance fluctuations in the test set. In a real problem the landscape of possibilities is unknown, which implies that the model fluctuations are, at least, unknown. This hints the difficulty in bounding the performance error for any unseen configuration. The naive solution is to increase the training data set. Increasing the training set arbitrarily will lead to an increase in training time. So care must be taken in such approach. One needs to take the best of both worlds, i.e., models trained on sufficiently large data sets to reduce fluctuations but small enough in order be able to train in a fashionable time. A better curated data set where configuration redundancy is kept at a minimum can lead to better performance. Another approach that seem promising is active learning \cite{sorscher2022beyond} whereby data is ranked by the magnitude of the performance error and data with the largest error is then fed into the training.

Defining the right metrics in deep learning is highly challenging. Partly because quantifying the degree of success can be difficult, whereas it is fairly easy agreeing in the ideal success. In other words, quantifying \textit{good enough} is not straightforward and requires bench-marking different approaches for comparison. A thorough discussion on benchmark suite can be found in \cite{thiyagalingam2022scientific}.

\section{Conclusions}


When selecting a NN for a specific task, it is important to consider the function and requirements of the task at hand. There is currently no consensus on which NN is optimal for a given task, primarily due to the large number of NN options available, the rapidly evolving nature of the field, and the lack of a comprehensive deep learning theory. This leads to a reliance on empirical results. Our paper is an important step at establishing best practices for this type of problems. we focused on randomly placed sources with random fluxes which yield large variations in the field. Our method can be generalized for different diffusion equations.

As part of future steps, we will increase the complexity of the problem being solved by considering conditions closer to real-problems, \textit{i.e.}, by considering less symmetrical sources, different diffusivities and different boundary conditions. In addition, further design is required for these models to be used in a production environment in a reliable way, \textit{i.e.}, how to deal with error performance edge cases on-the-fly? Ultimately, to be able to deploy the model for production, one requires a method to keep the performance error below a predefined bound. For instance, one can train an additional NN that takes the predicted stationary solution and predicts the initial condition, which is then compared with the ground truth initial condition. This framework allows a comparison between ground truth input and predicted input without requiring the ground truth steady-state solution. However, this approach would only be reliable if the input error is always proportional to the steady-state error and hence requires further investigation. A perhaps simpler approach consists in sampling the NN's output for the same input using a \textit{drop-out} feature, such that if the fluctuations from the samples are small enough, then the NN's prediction is robust enough. Both cases require benchmark design. We have developed a process in this paper that can easily be replicated for more complicated problems of this type, and provided a variety of benchmarks.

\section{Acknowledgements}

JQTM acknowledges a Mitacs Postdoctoral Fellowship. GCF acknowledges partial support from DOE DE-SC0023452 and NSF OAC-2204115.

\bibliographystyle{unsrt}  
\bibliography{references}

\end{document}